\pdfoutput=1
\documentclass[conference]{IEEEtran}
\usepackage{times}

\usepackage[numbers]{natbib}
\usepackage{multicol}
\usepackage[bookmarks=true]{hyperref}

\newcommand{\fig}[1]{Fig.~\ref{fig:#1}}
\newcommand{\tabl}[1]{Table~\ref{tab:#1}}
\newcommand{\sect}[1]{Sec.~\ref{sec:#1}}

\usepackage{microtype}
\usepackage{graphicx}
\usepackage{subfigure}
\usepackage{xcolor} 
\usepackage{booktabs} 
\usepackage{algorithm}
\usepackage{algorithmic}
\usepackage{amsfonts}
\usepackage{amsmath}
\usepackage{amssymb}
\usepackage{calrsfs}
\usepackage[font=footnotesize]{caption}
\usepackage{wrapfig}

\DeclareMathAlphabet{\mathcal}{OMS}{zplm}{m}{n}

\newcommand{\DPLAY}{D_{\operatorname{play}}}
\newcommand{\DPLAYLANG}{D_{\operatorname{(play,lang)}}}
\newcommand{\DDEMOLANG}{D_{\operatorname{(demo,lang)}}}
\newcommand{\senc}{s_{\operatorname{enc}}}
\newcommand{\genc}{g_{\operatorname{enc}}}

\newcommand{\lfpwebsite}{https://learning-from-play.github.io}
\newcommand{\lfpvideos}{\lfpwebsite/assets/mp4}
\newcommand{\website}{https://language-play.github.io}

\title{Language Conditioned Imitation Learning\\over Unstructured Data}

\author{Author Names Omitted for Anonymous Review. Paper-ID 168}

\author{
\authorblockN{Corey Lynch \IEEEauthorrefmark{1}}
\authorblockA{\href{http://g.co/robotics}{Robotics at Google}\\
\texttt{coreylynch@google.com}}
\and
\authorblockN{Pierre Sermanet \IEEEauthorrefmark{1}}
\authorblockA{\href{http://g.co/robotics}{Robotics at Google}\\
\texttt{sermanet@google.com}}
}

\begin{document}
\maketitle

\begingroup\renewcommand\thefootnote{}
\footnotetext{\IEEEauthorrefmark{1} Equal contribution}

\begin{abstract}
    Natural language is perhaps the most flexible and intuitive way for humans to communicate tasks to a robot.
    Prior work in imitation learning typically requires each task be specified with a task id or goal image---something that is often impractical in open-world environments.
    On the other hand, previous approaches in instruction following allow agent behavior to be guided by language, but typically assume structure in the observations, actuators, or language that limit their applicability to complex settings like robotics.
    In this work, we present a method for incorporating free-form natural language conditioning into imitation learning. Our approach learns perception from pixels, natural language understanding, and multitask continuous control end-to-end as a single neural network.
    Unlike prior work in imitation learning, our method is able to incorporate unlabeled and unstructured demonstration data (i.e. no task or language labels). We show this dramatically improves language conditioned performance, while reducing the cost of language annotation to less than 1\% of total data.
    At test time, a single language conditioned visuomotor policy trained with our method can perform a wide variety of robotic manipulation skills in a 3D environment, specified only with natural language descriptions of each task (e.g. ``open the drawer...now pick up the block...now press the green button...") (\href{\website/\#long-horizon}{see video}).
    To scale up the number of instructions an agent can follow, we propose combining text conditioned policies with large pretrained neural language models. We find this allows a policy to be robust to many out-of-distribution synonym instructions, without requiring new demonstrations.
    See videos of a human typing live text commands to our agent at \href{\website}{\website}

\end{abstract}

\IEEEpeerreviewmaketitle



\begin{figure*}[htb!]
  \centering
  \includegraphics[width=\linewidth]{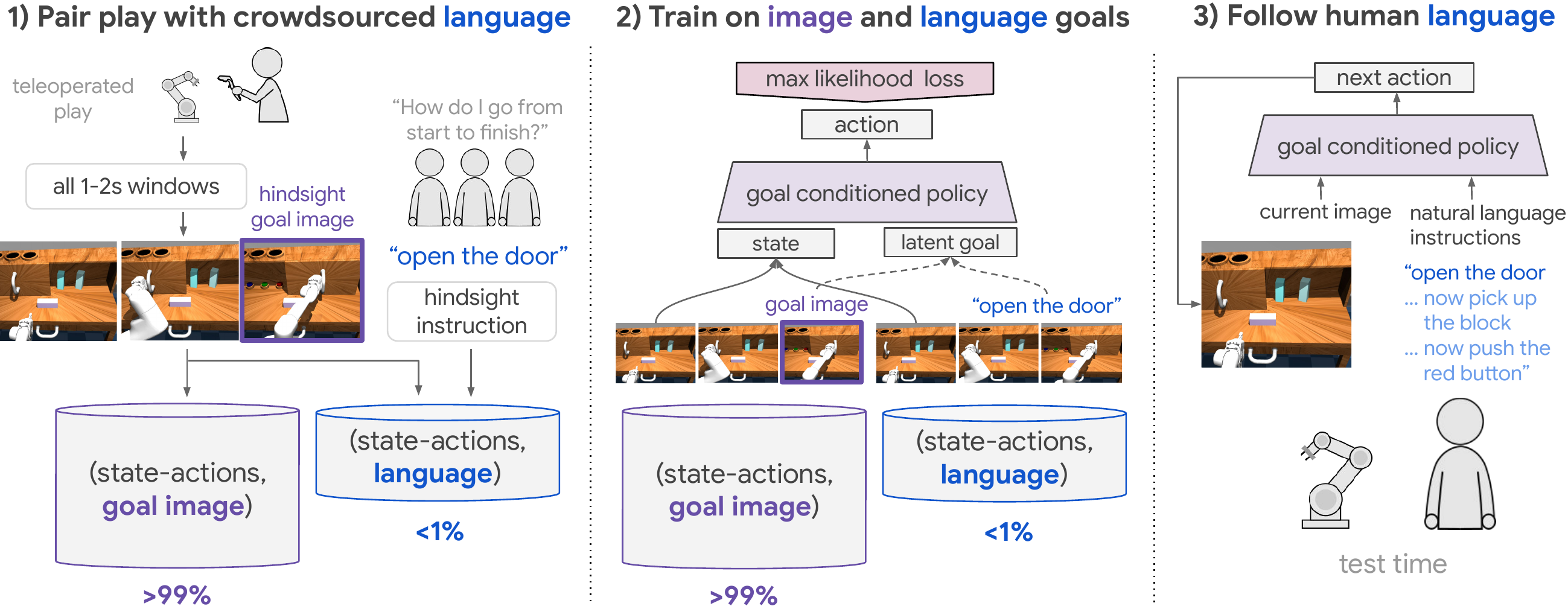}
  \caption{\textbf{Scaling up free-form natural language instruction following with unstructured data.} 1) First, relabel unstructured teleoperated ``play" data (no task labels) into hindsight goal image examples. Next, pair a small amount of play with hindsight instructions. 2) Train a single policy to follow either image or language goals. 3) Use only language conditioning at test time.}
  \label{fig:teaser}
\end{figure*}

\section{Introduction}
    Imitation learning \cite{billard2008survey, argall2009survey} is a popular framework for acquiring complex robotic skills from raw onboard sensors. 
    Traditionally, imitation learning has been applied to learning individual skills from structured and isolated human demonstrations \cite{schaal1999imitation, akgun2012keyframe, zhang2018deep}.

    These collection requirements are difficult to scale to real world scenarios, where robots are expected to be \textit{generalists}---capable of autonomously performing a wide variety of skills. As we consider deploying imitation learning in this setting, a critical challenge is scale: how can we lessen the data requirements of learning each new skill? Is it possible instead to learn many skills simultaneously from large amounts of unstructured, \textit{unlabeled} demonstration data \cite{hausman2017multi, lynch2019play, gupta2019relay}?
    
    Recent works have focused on scaling up multitask imitation learning over unstructured data \cite{ding2019goal, lynch2019play}. However, these approaches typically assume that tasks are specified to the agent at test time via mechanisms like one-hot task selectors \cite{rahmatizadeh2018vision, singh2020scalable}, goal images \cite{nair2018visual, florensa2019self, lynch2019play}, or target configurations of the state space \cite{gupta2019relay}. While these types of conditioning are straightforward to provide in simulation, they are often impractical to provide in open-world settings. Thus, another important consideration when deploying imitation learning in everyday settings is scalable \textit{task specification}: how can untrained users instruct robot behavior? This motivates robots that can follow free-form natural language instructions.
    
    Training agents to follow instructions is an active area of research in the broader machine learning community \cite{luketina2019survey}, yet it remains a difficult open challenge. Prior approaches have trained agents that map observations and language inputs directly to actions using neural networks, but often make assumptions that limit their applicability to robotics. Typical studies involve 2D observation spaces (e.g. games \cite{kollar2010toward,oh2017zero,misra2017mapping,1903.02020} and gridworlds \cite{yu2018interactive}), simplified actuators, (e.g. binary pick and place primitives \cite{hill2019emergent,wang2019reinforced}), or synthetic predefined language \cite{hermann2017grounded, chevalier2018babyai,1906.07343,zhong2019rtfm}.
    
    In this work, we study the setting of human language conditioned robotic manipulation (\fig{teaser}, step 3).
    In this setting, a single agent must execute a series of visual manipulation tasks in a row, each expressed in free-form natural language, e.g.  ``open the door all the way to the right...now grab the block...now push the red button...now open the drawer". Furthermore, agents in this scenario are expected to be able to perform any combination of subtasks in any order.
    This is the first version of instruction following, to our knowledge, that combines: natural language conditioning, high-dimensional pixel inputs, 8-DOF continuous control, and complex tasks like long-horizon robotic object manipulation.
    Text conditioning is also a new and difficult setting for goal-directed imitation learning \cite{lynch2019play, ding2019goal}, which introduces important new research questions. For example, how can we learn the mapping between language and actions with the fewest language labels? How can we leverage large unstructured demonstration datasets that have no language labels? How can we follow the maximum number of instructions at test time, given a finite training set?
    
    To address this setting, we propose a simple approach for combining imitation learning with free-form text conditioning (\fig{teaser}).
    Our method consists of only standard supervised learning subroutines, and learns perception, language understanding, and control end-to-end as a single neural network.
    Critically, unlike prior work in instruction following, our method can leverage large amounts of unstructured and unlabeled demonstration data (i.e. with no language or task labels). We show this reduces the burden of language annotation to less than 1\% of total data.
    To scale up the maximum amount of instructions our agent can follow at test time, we introduce a simple technique for combining any language conditioned policy with \textit{large pretrained language models} \cite{devlin2018bert, radford2019language}. We find that this simple modification allows our agent to follow thousands of new synonym instructions at test time, without requiring that new robot demonstrations be collected for each synonym.
    We believe that the capabilities proposed here of learning from unlabeled demonstrations, end-to-end learning of text conditioned visuomotor policies, and following new synonym instructions without new robot data constitute important steps towards scalable robot learning systems.
    
    We evaluate our method in a dynamically accurate simulated 3D tabletop environment with a fixed set of objects. Our experiments show that a language conditioned visuomotor policy trained with our method can perform many complex robotic manipulation skills in a row specified entirely with natural language (see \href{\website/videos/live/playlang_20200326-193259_13tasks.webm}{video}), outperforming conventional imitation baselines trained on structured data.

    \textbf{Contributions} To summarize the contributions of this work, we:
    \begin{itemize}
        \item introduced a setting of human language conditioned robotic visual manipulation.
        \item introduced a simple learning method for combining free-form text conditioning with multitask imitation learning.
        \item introduced \textit{multicontext imitation learning}, applicable to any contextual imitation learning setup, allowing us to train a language conditioned policy over mostly unstructured and unlabeled demonstrations (i.e. with no task or language labels).
        \item demonstrated that the resulting language conditioned visuomotor policy can follow many free-form human text instructions over a long horizon in a simulated 3D tabletop setting, i.e. “open the door…now pick up the block…now press the red button” (see video).
        \item introduced a simple way to combine any language conditioned policy with large pretrained language models. We show this improves manipulation performance and allows an agent to be robust to thousands of new synonym instructions at test time, in 16 languages, without requiring new demonstrations for each synonym.
    \end{itemize}

\begin{figure*}[htb]
\begin{minipage}[c]{0.43\linewidth}
        \begin{center}
        \includegraphics[width=1\linewidth]{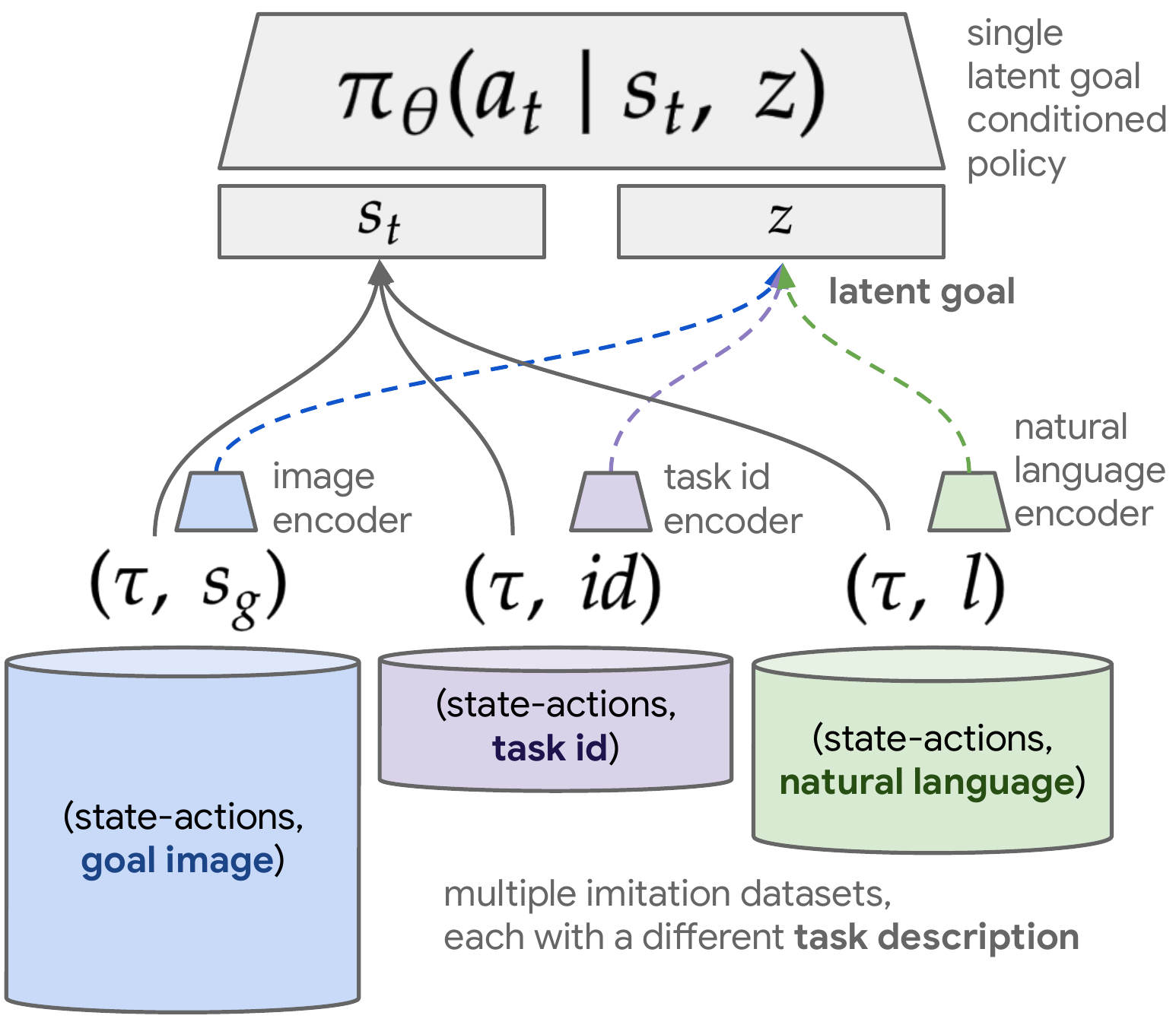}
        \caption{\textbf{Multicontext Imitation Learning (MCIL).} We introduce a simple generalization of contextual imitation learning to multiple heterogeneous contexts (e.g. goal image, task id, natural language). MCIL trains a single \textit{latent goal} conditioned policy over all datasets simultaneously, as well as one encoder per dataset, each mapping to the shared latent goal space. This allows training a language conditioned policy over both labeled and unlabeled demonstration datasets.}
        \label{fig:acgc}
        \end{center}
\end{minipage}
\hspace{0.01\linewidth}
\begin{minipage}[c]{0.55\linewidth}
        \begin{center}
        \centerline{\includegraphics[width=.65\linewidth]{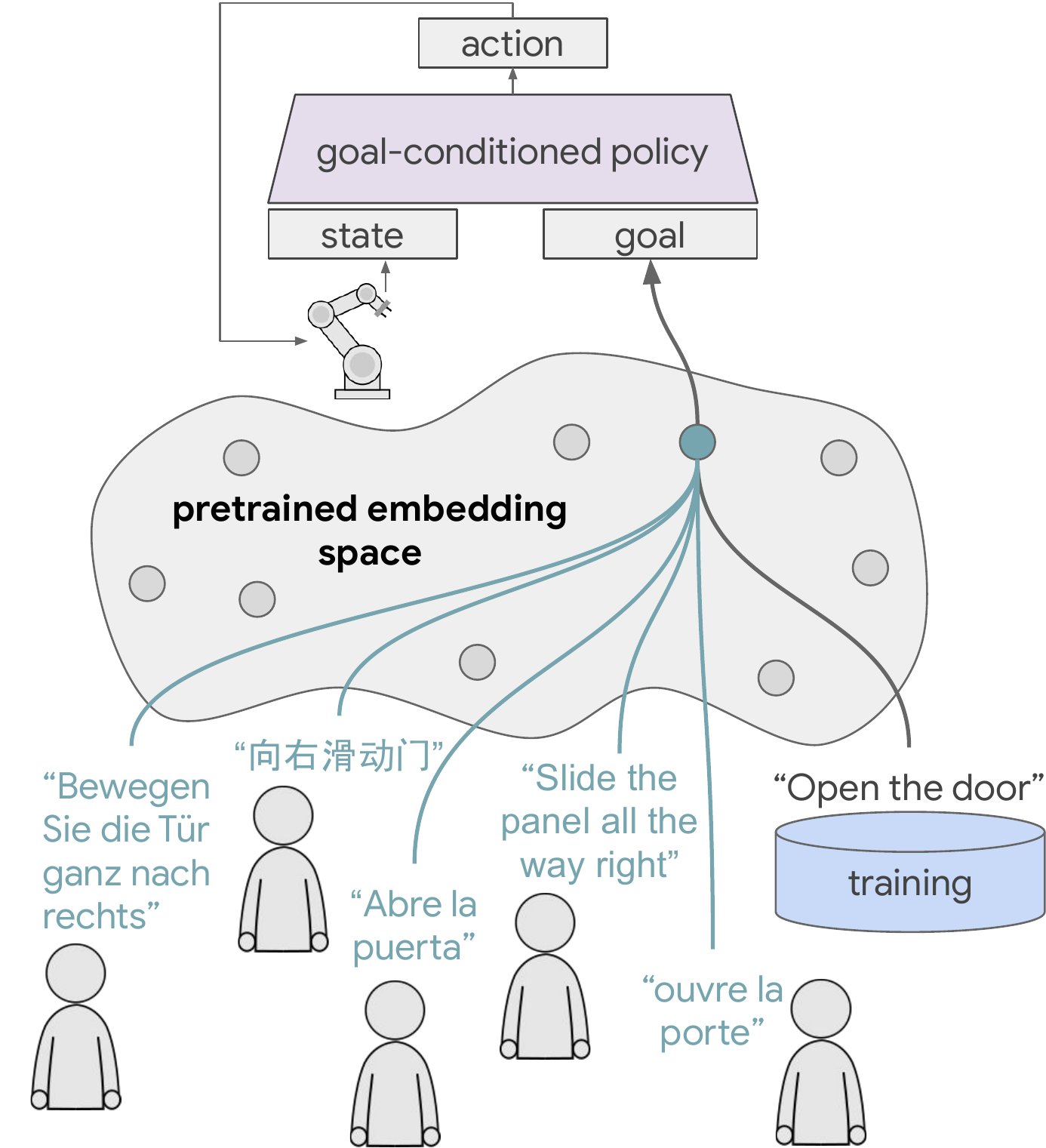}}
        \caption{\textbf{Pretrained language models make language conditioned policies robust to out-of-distribution synonyms}.
        Simply by training on top of pretrained language embeddings, we can give a language conditioned policy the ability to follow out-of-distribution synonym instructions at test time. The pretrained embedding space is responsible for relating new synonym instructions (green) to ones from the agent's training set (black).
        }
      \label{fig:follow_ood}
        \end{center}
\end{minipage}%
\end{figure*}

\section{Related Work}
    \textbf{Robotic learning from general sensors}. 
    In general, learning complex robotic skills from low-level sensors is possible, but requires substantial human supervision. Two common approaches are imitation learning (IL) \cite{argall2009survey} and reinforcement learning (RL) \cite{kober2013reinforcement}. When combined with deep function approximators, IL typically requires many human demonstrations \cite{rajeswaran2017learning,rahmatizadeh2018vision,zhang2018deep} to drive supervised learning of a policy. In RL, supervision takes the form of hand-designed task rewards. Reward design is non-trivial in complex environments, often requiring either task-specific instrumentation \cite{gu2017deep} or learned perceptual rewards \cite{Sermanet2017TCN,singh2019end}. Additionally, RL agents often require hand-designed strategies \cite{kalashnikov2018qt,ebert2018visual} or human demonstrations \cite{rajeswaran2017learning} to overcome hard exploration problems. Finally, even under multitask formulations of RL \cite{teh2017distral} and IL \cite{rahmatizadeh2018vision}, each new task considered requires a corresponding and sizable human effort. This makes it difficult to scale either approach naively to a broad task setting. In this work, we focus on scaling imitation learning to be multitask and language conditioned. While there are several ways to perform imitation learning such as inverse reinforcement learning \cite{ng2000algorithms, wulfmeier2015maximum, finn2016guided} and occupancy matching \cite{ho2016generative}, we restrict our attention in this work to behavior cloning \cite{pomerleau1991efficient} given its stability and ease of use.
    
    \textbf{Imitation learning from large unstructured datasets}.
    Recent works \cite{hausman2017multi, lynch2019play, gupta2019relay} have sought to mitigate the costs of conventional multitask imitation learning by instead learning many skills at once over large unstructured demonstration datasets. Like these works, we incorporate unstructured demonstration data into our method. Unlike these works, the resulting goal conditioned policies can be conditioned with natural language. 
    
    \textbf{Task agnostic control}. This paper builds on the setting of task agnostic control, where a single agent is trained to reach any reachable goal state in its environment upon command \cite{kaelbling1993learning, schaul2015universal}. One way of acquiring this kind of control is to first learn a model of the environment through interaction \cite{oh2015action,ebert2018visual} then use it for planning. However,
    these approaches rely on accurate forward models of visual dynamics, a challenging open problem. A powerful model-free strategy for task agnostic control is goal relabeling \cite{kaelbling1993learning,andrychowicz2017hindsight}. This technique trains goal conditioned policies to reach any previously visited state upon demand, with many recent examples in RL \cite{nair2018visual,gupta2019relay,nair2020contextual} and IL \cite{lynch2019play,ding2019goal}. A limitation to models combining relabeling with image observation spaces is that tasks must be specified with goal images at test time. The present work builds on relabeled imitation, but additionally equips policies with natural language conditioning.

    \textbf{Multicontext learning}. A number of previous methods have focused on generalizing across tasks \cite{caruana1997multitask}, or generalizing across goals \cite{schaul2015universal}. We introduce multicontext learning, a framework for generalizing across heterogeneous task and goal descriptions. When one of the training sources is plentiful and the other scarce, multicontext learning can be seen as transfer learning \cite{tan2018survey} through a shared goal space. Multicontext imitation is a central component of our method, as it reduces the cost of human language supervision to the point where it can be practically applied.

    \textbf{Instruction following}. There is a long history of research into agents that not only learn a grounded language understanding \cite{winograd1972understanding}, but demonstrate that understanding by following instructions (survey \cite{luketina2019survey}). Recently, authors have had success using deep learning to directly map raw input and text instructions to actions. However, prior work has often studied 2D environments \cite{kollar2010toward,oh2017zero,misra2017mapping,yu2018interactive} and simplified actuators \cite{wang2019reinforced,hill2019emergent,1903.02020}. Additionally, learning to follow \textit{natural} language is still not the standard in instruction following research \cite{luketina2019survey}, with typical implementations instead assuming access to simulator-provided instructions drawn from a restricted vocabulary and grammar \cite{hermann2017grounded}.
    This work, in contrast, studies 1) natural language instructions, 2) high-dimensional continuous sensory inputs and actuators, and 3) complex tasks like long-horizon 3D robotic object manipulation. Furthermore, unlike existing RL approaches to instruction following \cite{hill2019emergent}, our IL method is sample efficient, requires no reward definition, and scales easily to the multitask setting. Unlike concurrent IL approaches to robotic instruction following \cite{stepputtis2020language}, which assume access to labeled task demonstrations and pretrained object detection, our method learns from unstructured and unlabeled demonstration data and learns perception, language understanding, and control fully end-to-end via a single imitation loss.

\section{Problem Formulation}
We consider the problem of learning a natural language conditioned control policy $\pi_{\theta}(a | s, l)$, which outputs next action $a \in A$, conditioned on current state $s \in S$ and free-form natural language $l \in L$ describing a short-horizon task. Note that $S$ is not the true environment state, but rather the high dimensional onboard observations of the robot, e.g. $S = \{\operatorname{image}, \operatorname{proprioception}\}$. $l$ is provided by humans and has no limits on vocabulary or grammar.

At test time, a human gives a robot a series of instructions in a row, one at a time: $\{l_0, ..., l_N\}$. The language conditioned visuomotor policy $\pi_{\theta}(a | s, l)$ issues high frequency continuous control in closed loop to obtain the desired behavior described in $l$. The human may provide new instructions $l$ to the agent at any time, either commanding a new subtask or providing guidance (i.e. ``move your hand back slightly”).

Standard imitation learning setups learn single task policies $\pi_{\theta}(a | s)$ using supervised learning over a dataset $\mathcal{D} = \{\tau_i\}^N_i$, of expert state-action trajectories $\tau = \{(s_0, a_0), ... \}$. Closest to our problem setting is goal conditioned imitation learning \cite{kaelbling1993learning}, which instead aims to learn a policy $\pi_{\theta}(a | s, g)$, conditioned on $s \in S$ and a task descriptor $g \in G$ (often a one-hot task encoding \cite{rahmatizadeh2018vision}). When tasks can be described as a goal state $g = s_g \in S$ to reach, this allows any state visited during collection to be \textit{relabeled} \cite{kaelbling1993learning,andrychowicz2017hindsight,ding2019goal} as a ``reached goal state", with the preceding states and actions treated as optimal behavior for reaching that goal. At test time, learned behaviors are conditioned using a goal image $s_g$.

Relabeled imitation learning can be applied to unstructured and unlabeled demonstration data \cite{lynch2019play} to learn general purpose goal-reaching policies. Here, demonstrators are not constrained to a predefined set of tasks, but rather engage in every available object manipulation in a scene (\href{\lfpvideos/play_data516x360.mp4}{see example}). This yields one long unsegmented demonstration of semantically meaningful behaviors, which can be relabeled using Algorithm~\ref{alg:collect_play} into a training set $\DPLAY = \{(\tau, s_g)_i\}^{\DPLAY}_{i=0}$. These short horizon goal image conditioned demonstrations are fed to a maximum likelihood goal conditioned imitation objective: $\mathcal{L}_{\operatorname{GCIL}} = \mathbb{E}_{(\tau,\ s_g) \sim \DPLAY}\left[\sum ^{|\tau|}_{t=0}\operatorname{log} \pi _{\theta}\left( a_{t}| s_t, s_g\right)\right]$.

However, when learning language conditioned policies $\pi_{\theta}(a | s, l)$, we cannot easily relabel any visited state $s$ to a natural language goal as the goal space $G$ is no longer equivalent to the observation space $L$. Consequently, this limits language conditioned imitation learning from being able to incorporate large unstructured demonstration datasets $\DPLAY$.

Next, we describe our approach that relaxes this limitation.

\section{Learning to Follow Human Instructions from Unstructured Data}
    
    \label{sec:main_method}
    
    
    We present a framework that aims for a scalable combination of self-supervised relabeled imitation learning and labeled instruction following. Our approach (\fig{teaser}, Sec.~\ref{sec:lang_lfp}) can be summarized as follows:
    \begin{enumerate}
        \item Collection: Collect a large unstructured ``play" demonstration dataset. 
        \item Relabeling: Relabel unstructured data into goal image demonstrations. Pair a small number of random windows with language after-the-fact. (Sec.~\ref{sec:hip})
        \item Multicontext imitation: Train a single imitation policy to solve for either goal image or language goals. (Sec.~\ref{sec:acgc})
        \item Use only language conditioning at test time.
    \end{enumerate}

    \vspace{-.05in}
    \subsection{Pairing unstructured demonstrations with natural language}
    \vspace{-.05in}
    \label{sec:hip}
    Learning to follow language instructions involves addressing a difficult language grounding problem \cite{harnad1990symbol}: how do agents relate unstructured language $l$ to their onboard perceptions $s$ and actions $a$? We take a statistical machine learning approach, creating a paired corpora of (demonstration, language) by pairing random windows from unstructured data $\DPLAY$ with \textit{hindsight instructions} after-the-fact (Algorithm~\ref{alg:collect_play_lang}, part 1 of Fig.~\ref{fig:teaser}).
    Here, we show untrained human annotators on-board videos, then ask them ``what instruction would you give the agent to get from first frame to last frame?" See training examples in Table~\ref{tab:ex_train_lang}, video examples \href{\website/\#data}{here}, and a full description of the collection process in Appendix~\ref{sec:appendix_dataset_hip}. Annotators were encouraged to use free-form natural language and not constrain themselves to predefined vocabulary or grammar. This yields a new dataset $\DPLAYLANG = \{(\tau, l)_i\}^{\DPLAYLANG}_{i=0}$, a dataset of text conditioned demonstrations. Crucially, we do not need pair every window from play with language to learn to follow instructions. This is made possible with Multicontext Imitation Learning, described next. 

        \begin{algorithm}[htb]
        {\small
          \caption{Multicontext imitation learning}
          \label{alg:train_acgc}
          \begin{algorithmic}[1]
            \STATE {\bfseries Input:} $\mathcal{D} = \{D^0, ...,\ D^K\}, D^k = \{(\tau^{k}_i, c_{i}^k)\}^{D^k}_{i=0}$, One dataset per context type (e.g. goal image, language instruction, task id), each holding pairs of (demonstration, context).
            \STATE {\bfseries Input:} $\mathcal{F} = \{f_{\theta}^0, ...,\ f_{\theta}^K\}$, One encoder per context type, mapping context to shared latent goal space, i.e. $z = f_{\theta}^k(c^k)$.
            \STATE {\bfseries Input:} $\pi_{\theta}(a_t | s_t, z)$, Single latent goal conditioned policy.
            \STATE {\bfseries Input:} Randomly initialize parameters $\theta = \{\theta_{\pi},\ \theta_{f^0},\ ...,\ \theta_{f^K}\}$
            \WHILE{True}
              \STATE $\mathcal{L}_{\operatorname{MCIL}} \xleftarrow{} 0$
              \STATE {\color{gray} \# Loop over datasets.}
              \FOR {$k=0 ... K $}
                \STATE {\color{gray} \# Sample a (demonstration, context) batch from this dataset.}
                \STATE $(\tau^k, c^k) \thicksim D^k$
                \STATE {\color{gray} \# Encode context in shared latent goal space.}
                \STATE $z = f_{\theta}^k(c^k)$
                \STATE {\color{gray} \# Accumulate imitation loss.}
                \STATE $\mathcal{L}_{\operatorname{MCIL}} \mathrel{{+}{=}} \sum^{|\tau^k|}_{t=0}\operatorname{log} \pi _{\theta}\left( a_{t}| s_t, z\right)$
              \ENDFOR
              \STATE {\color{gray} \# Average gradients over context types.}
              \STATE $\mathcal{L}_{\operatorname{MCIL}} \mathrel{{*}{=}} \dfrac{1}{|\mathcal{D}|}$
              \STATE {\color{gray} \# Train policy and all encoders end-to-end.}
              \STATE Update $\theta$ by taking a gradient step w.r.t. $\mathcal{L}_{\operatorname{MCIL}}$
            \ENDWHILE
          \end{algorithmic}
        }
    \end{algorithm}
        
    \subsection{Multicontext Imitation Learning}
    \label{sec:acgc}
    
    So far, we have described a way to create two contextual imitation datasets: $\DPLAY$, holding goal image demonstrations and $\DPLAYLANG$, holding language demonstrations. 
    Ideally, we could train a single imitation policy that could be conditioned with \textit{either} task description. 
    Critically, this would enable language conditioned imitation to make use of self-supervised imitation over unstructured data.

    With this motivation, we introduce \emph{multicontext imitation learning} (MCIL), a simple and universally applicable generalization of contextual imitation to multiple heterogeneous contexts. The main idea is to represent a large set of policies by a single, unified function approximator that generalizes over states, tasks, and \textit{task descriptions}. Concretely, MCIL assumes access to multiple contextual imitation datasets $\mathcal{D} = \{D^0, ...,\ D^K\}$, each with a different way of describing tasks. Each $D^k = \{(\tau^{k}_i, c_{i}^k)\}^{D^k}_{i=0}$ holds demonstrations $\tau$ paired with some context $c \in C^k$. For example, $D^0$ might contain one-hot task demonstrations (a conventional multitask imitation dataset), $D^1$ might contain goal image demonstrations (obtained by hindsight relabeling), and $D^2$ might contain language goal demonstrations. MCIL trains a single \emph{latent goal} conditioned policy $\pi_{\theta}(a_t | s_t, z)$ over \textit{all} datasets simultaneously, as well as one parameterized encoder per dataset $\mathcal{F} = \{f_{\theta}^0, ..., f_{\theta}^K\}$, learning to map raw task descriptions to a common latent goal space $z \in \mathbb{R}^d$. See Fig.~\ref{fig:acgc}. For instance, these could be a one-hot task embedding lookup, an image encoder, and a language encoder respectively.
    
    MCIL has a simple training procedure, shown in Algorithm~\ref{alg:train_acgc}: at each training step, sample a batch of contextual imitation examples from each dataset, encode contexts in the shared latent space using the respective encoders, then compute a latent goal-conditioned imitation loss, averaged over all datasets. The policy and goal encoders are trained end-to-end to maximize this objective. 
    
    MCIL allows for a highly efficient training scheme, broadly useful beyond this paper: learn the majority of control from the data source that is cheapest to collect, while simultaneously learning scalable task conditioning from a small number of labeled examples. As we will see empirically, MCIL allows us to train an instruction following agent with less than 1\% of data requiring language annotation, with the majority of perception and control learned instead from relabeled goal image imitation.

    \begin{figure*}[htb]
      \centering
      \includegraphics[width=.9\linewidth]{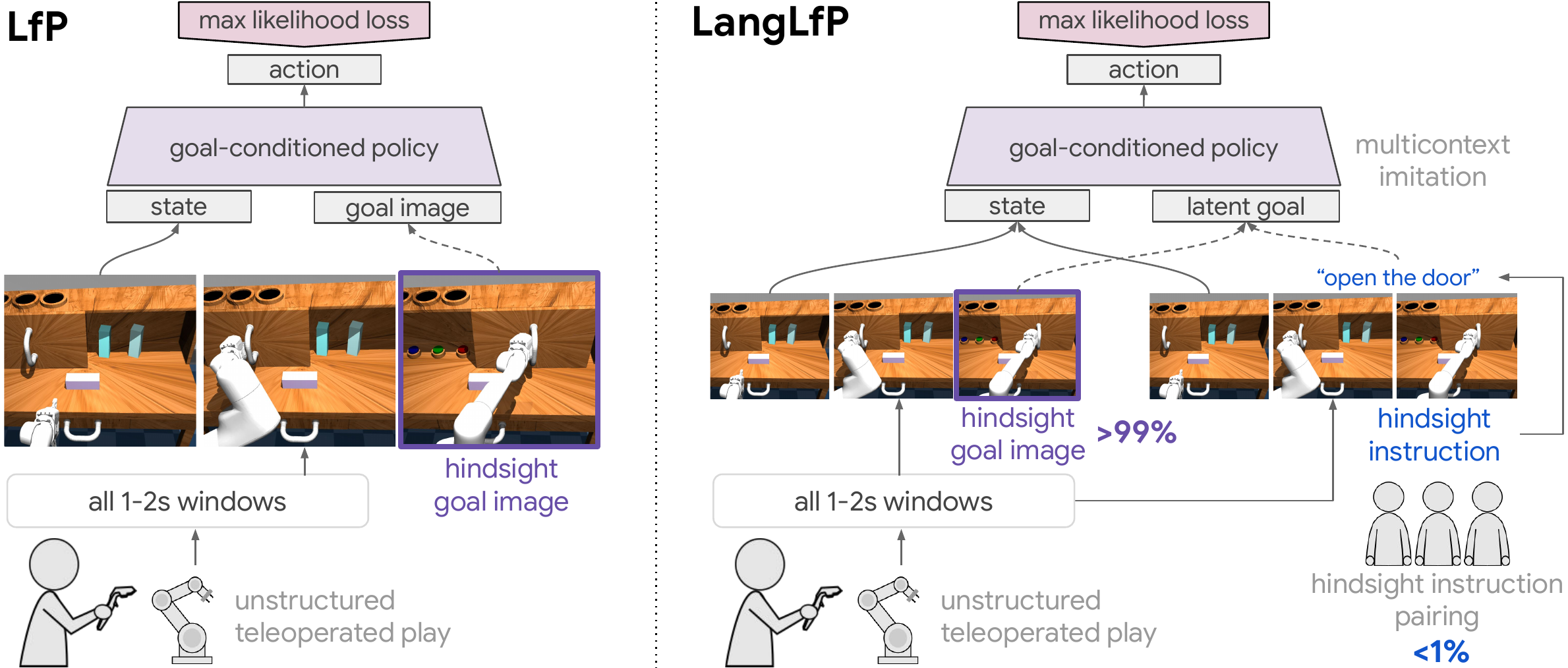}
      \caption{\textbf{Comparing prior LfP (left) to our LangLfP (right)}. Both are trained on unstructured and unsegmented demonstrations, relabeled into goal image demonstrations. LangLfP is additionally trained on random windows paired with hindsight natural language instructions. 
    }
      \label{fig:comparing_lfp_to_langlfp}
    \end{figure*}

    \subsection{LangLfP: following image or language goals.}
    \label{sec:lang_lfp}
    We now have all the components to introduce \emph{LangLfP} (language conditioned learning from play). LangLfP is a special case of MCIL (Sec.~\ref{sec:acgc}) applied to our problem setting, and learns perception from pixels, natural language understanding, and control end-to-end with no auxiliary losses.

    \textbf{Training LangLfP.} LangLfP trains a single multicontext policy $\pi_\theta(a_t | s_t, z)$ over datasets $\mathcal{D} = \{\DPLAY, \DPLAYLANG\}$. We define $\mathcal{F} = \{\genc, \senc\}$, neural network encoders mapping from goal images (Appendix~\ref{sec:appendix_genc}) and text instructions (Appendix~\ref{sec:appendix_nlu}) respectively to $z$. Fig.~\ref{fig:comparing_lfp_to_langlfp} compares LangLfP training to prior LfP training. 
    At each training step, LangLfP: 1) samples a batch of image goal tasks from $\DPLAY$ and a batch of language goal tasks from $\DPLAYLANG$, 2) encodes image and language goals into $z$, and 3) computes the MCIL objective. We then take a combined gradient step with respect to all modules---perception (Appendix~\ref{sec:appendix_perception}), language (Appendix~\ref{sec:appendix_nlu}), and control (Appendix~\ref{sec:appendix_control})---optimizing the whole architecture end-to-end as a single neural network. See full training details in Appendix~\ref{sec:appendix_training}.
    
    \textbf{Following natural language instructions at  test time.} At test time, LangLfP conditions only on onboard pixel observations and a free-form natural language task description to solve user-defined tasks in closed loop. See picture in part 3 of Fig.~\ref{fig:teaser} and details in Appendix~\ref{sec:appendix_test}.

    \textbf{Leveraging pretrained language models}
    While LangLfP offers a straightforward way for learning natural language end-to-end from imitation, this may not always be desirable. In open-world scenarios, instruction following robots are likely to be given instructions that are \textit{synonyms} to ones they have been trained to follow, but do not overlap exactly with a finite training set. For example, ``\textit{shut} the door" is a valid, but potentially out-of-distribution way of describing training task ``\textit{close} the door". Many recent works have successfully transferred knowledge from unlabeled text to downstream NLP tasks via pretrained embeddings \cite{devlin2018bert,radford2019language}. Can we achieve similar knowledge transfer to \textit{robotic manipulation}? There are two motivations for this kind of transfer: 1) improve language conditioned manipulation and 2) allow an agent to follow out of distribution synonym instructions (Fig.~\ref{fig:follow_ood}). To test these hypotheses, we augment LangLfP, encoding language inputs $l$ to the policy at training and test time in the pretrained embedding space of a multilingual neural language model \cite{yang2019multilingual}. We refer to this augmented model as \emph{TransferLangLfP}. See Appendix~\ref{sec:appendix_nlu} for details.

\section{Experiments}
Our experiments aim to answer the following questions:
Q0) Can a policy trained with our method learn perception, natural language understanding, and control end-to-end to solve many free-form text conditioned tasks in a 3D tabletop environment?
Q1) Can our language conditioned multitask imitation learning (LangLfP) match the performance of goal image conditioned imitation learning (LfP)?
Q2) Does our method's ability to leverage large unlabeled imitation datasets improve language conditioned performance?
Q3) Does incorporating pretrained language models grant any benefit to manipulation performance?
Q4) Does incorporating pretrained language models allow our agent to be robust to out of distribution synonym instructions?


\textbf{Dataset and tasks.} We conduct our experiments in the simulated 3D Playroom environment introduced in \cite{lynch2019play}, shown in Fig.~\ref{fig:env}.  It consists of an 8-DOF robotic arm controlled with continuous position control from visual observations at 30Hz
to accomplish manipulation tasks from 18 distinct task families. See \cite{lynch2019play} for a complete discussion of tasks. We modify the environment to include a text channel which the agent observes at each timestep, allowing humans to type unconstrained language commands (details in Appendix~\ref{sec:appendix_env}). We define two sets of experiments for each baseline: \textbf{pixel experiments}, where models receive pixel observations and must learn perception end-to-end, and \textbf{state experiments}, where models instead receive simulator state consisting of positions and orientations for all objects in the scene. The latter provides an upper bound on how all well the various methods can learn language conditioned control, independent of a difficult perception problem (which might be improved upon independently with self-supervised representation learning methods (e.g. \cite{Sermanet2017TCN,ozair2019wasserstein,pirk2019online})).

\textbf{Methods.} We compare the following methods (details on each in Appendix~\ref{sec:appendix_models}):
    \textbf{LangBC} (``language, but no play"): a baseline natural language conditioned multitask imitation policy \cite{rahmatizadeh2018vision}, trained on $\DDEMOLANG$, 100 expert demonstrations for each of the 18 evaluation tasks paired with hindsight instructions.
    \textbf{LfP} (``play, but no language"): a baseline LfP model trained on $\DPLAY$, conditioned on goal images at test time.
    \textbf{LangLfP (ours)} (``play and language"): Multicontext imitation trained on unstructured data $\DPLAY$ and unstructured data paired with language $\DPLAYLANG$. Tasks are specified at test time using only natural language.
    \textbf{Restricted LangLfP}: LangLfP trained on ``restricted $\DPLAY$", a play dataset restricted to be the same size as $\DDEMOLANG$. This restriction is somewhat artificial, as more unsegmented play can be collected for the same budget of time. This provides a controlled comparison to LangBC.
    \textbf{TransferLangLfP (ours)}: LangLfP trained on top of pretrained language embeddings.
To perform a controlled comparison, we use the same network architecture (details in \sect{appendix_architecture}) across all baselines. See Appendix~\ref{sec:appendix_dataset} for a detailed description of all data sources.

\subsection{Human Language Conditioned Visual Manipulation}

We construct a large number of multi-stage human language conditioned manipulation tasks by treating the original 18 evaluation tasks in \cite{lynch2019play} as subtasks, then considering all valid N-stage transitions between them. This results in 2-stage, 3-stage, and 4-stage benchmarks, referred to here as Chain-2, Chain-3, and Chain-4. See Appendix~\ref{sec:appendix_eval} for benchmark details. We obtain language instructions for each subtask in the same way as training (\sect{hip}), by presenting human annotators with videos of the completed tasks and asking for hindsight instructions. We present results in Table~\ref{tab:main_results} and Fig.~\ref{fig:long_horizon} and discuss below.

    \begin{table*}
    {\small
    \centering
    \begin{tabular}{|l|c|c|c|c|c|}
    \multicolumn{1}{|c|}{\textbf{Method}} & \textbf{Input} & \textbf{\begin{tabular}[c]{@{}c@{}}Training\\ source\end{tabular}} & \textbf{\begin{tabular}[c]{@{}c@{}}Task\\condi-\\tioning\end{tabular}} & \multicolumn{1}{c|}{\textbf{\begin{tabular}[c]{@{}c@{}}Multi-18\\Success\\(18 tasks)\end{tabular}}} & \multicolumn{1}{c|}{\textbf{\begin{tabular}[c]{@{}c@{}}Chain-4 Success\\ (925\\long-horizon tasks)\end{tabular}}}\\
    \hline
    \hline
    LangBC              & pixels & predefined demos & text & 20.0\% $\pm 3.0$ & 7.1\% $\pm 1.5$ \\
    Restricted LangLfP  & pixels & unstructured demos & text & 47.1\% $\pm 2.0$ & 25.0\% $\pm 2.0$ \\
    LfP                 & pixels & unstructured demos & image & 66.4\% $\pm 2.2$ & 53.0\% $\pm 5.0$ \\
    LangLfP (ours)      & pixels & unstructured demos & text & 68.6\% $\pm 1.7$ & 52.1\% $\pm 2.0$ \\
    TransferLangLfP (ours) & pixels & \textbf{unstructured demos} & \textbf{text} & \textbf{74.1\%} $\pm 1.5$ & \textbf{61.8}\% $\pm 1.1$ \\
    \hline
    LangBC              & states & predefined demos & text & 38.5\% $\pm 6.3$ & 13.9\% $\pm1.4$ \\
    Restricted LangLfP  & states & unstructured demos & text & 88.0\% $\pm 1.4$ & 64.2\% $\pm 1.5$ \\
    LangLfP (ours)      & states & unstructured demos & text & 88.5\% $\pm 2.9$ & 63.2\% $\pm 0.9$ \\
    TransferLangLfP (ours) & states & \textbf{unstructured demos} & \textbf{text} & \textbf{90.5\%} $\pm 0.8$ & \textbf{71.8\%} $\pm 1.6$
    \end{tabular}
    \caption{\textbf{Human language conditioned visual manipulation experiments}}
    \label{tab:main_results}
    }
    \end{table*}

    \begin{figure*}[tb]
        \begin{minipage}[c]{0.5\linewidth}
                \begin{center}
                \centerline{\includegraphics[width=.9\linewidth]{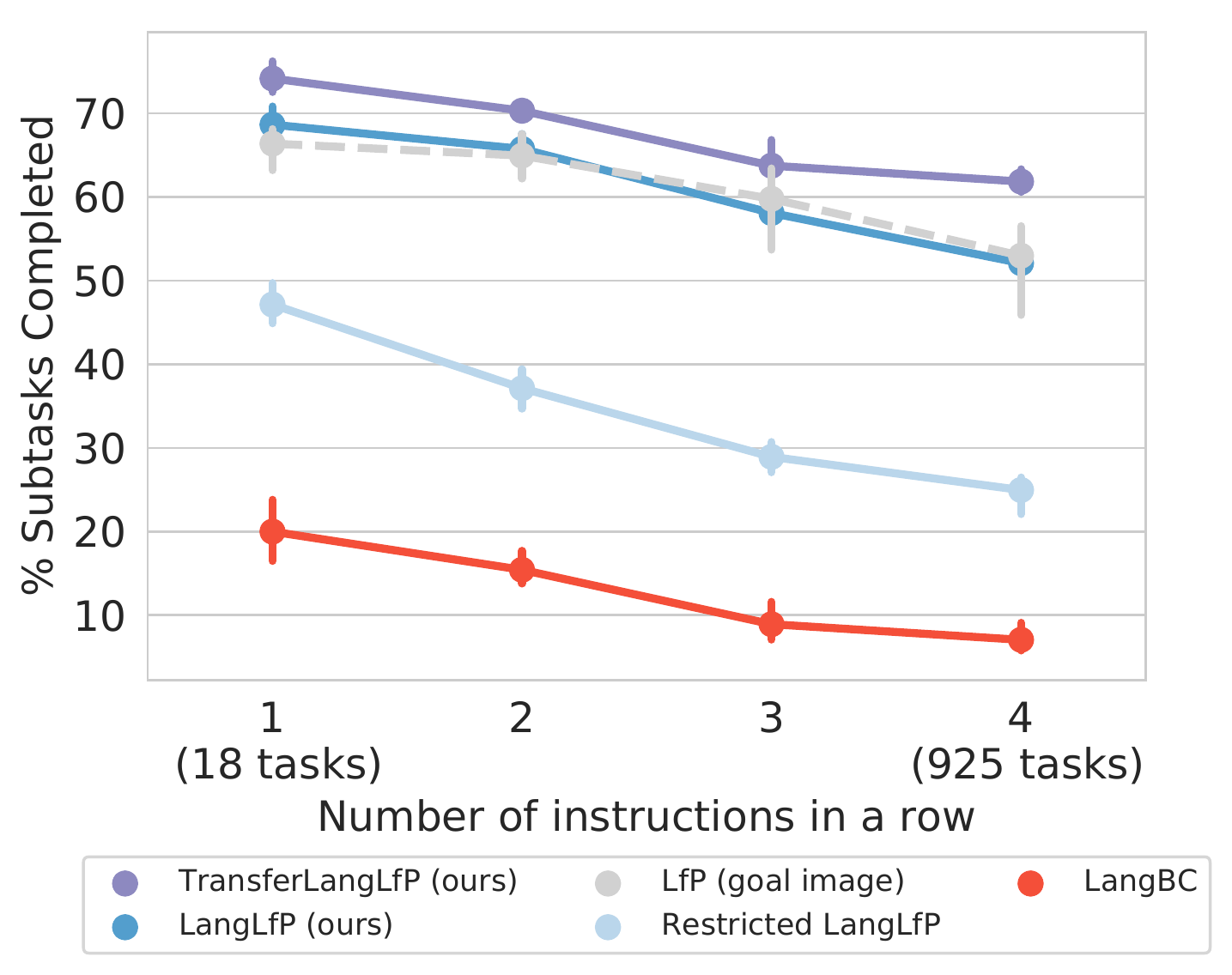}}
                \caption{\textbf{Long horizon language conditioned visual manipulation results.}}
                \label{fig:long_horizon}
                \end{center}
        \end{minipage}
        \begin{minipage}[c]{0.5\linewidth}
                \begin{center}
                \centerline{\includegraphics[width=.9\linewidth]{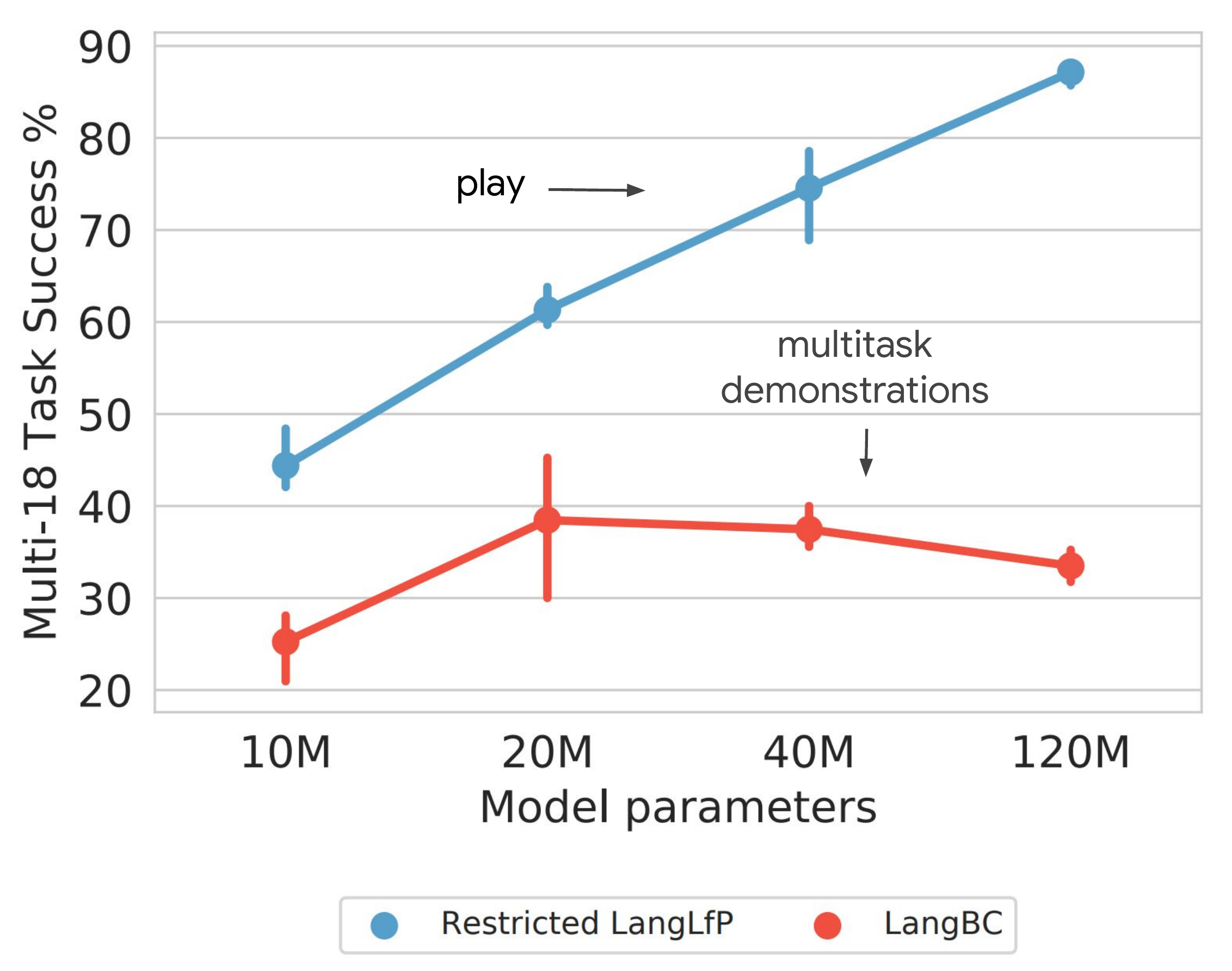}}
                \caption{\textbf{Performance scales linearly with model capacity, but only for unstructured demonstrations.} We see that for the same amount of data, more diverse unstructured imitation data is better utilized by large model learning.}
                \label{fig:sweep_capacity}
                \end{center}
        \end{minipage}%
    \end{figure*}
    
    \textbf{Language conditioned manipulation results}. In Table~\ref{tab:main_results}, we see that LangLfP can achieve 68.6\% success on individual free-form instructions (covering 18 different tasks), 52.1\% success on 925 4-stage instructions. This helps answer the first main question of this work (Q0), whether our approach can learn perception, control, and language understanding end-to-end to follow many free-form human instructions. We also see that LangLfP matches the performance of prior goal image conditioned LfP on all benchmarks within margin of error (Q1). This is important, as it shows a more scalable mode of task conditioning can be achieved with only $\thicksim$0.1\% of demonstration data requiring language annotation.
    
    \textbf{Large unlabeled imitation datasets improve language conditioned performance}.
    We see in Table~\ref{tab:main_results} and \fig{long_horizon} that LangLfP outperforms LangBC on every benchmark. This is important because it shows that by allowing our policy to train over unlabeled demonstration data (via MCIL), it achieves significantly better performance than baseline policies trained only on typical predefined task demonstrations (Q2). We note this holds even when unlabeled training sources are restricted to the same size as labeled ones (Restricted LangLfP vs. LangBC). Qualitatively (\href{\website/\#lfp-vs-bc}{videos}), we see clear differences between models that incorporate unstructured data and those that do not. We find that on long-horizon evaluations, MCIL-trained policies tend to transition well between tasks and recover from initial failures, whereas baseline policies trained on conventional demonstrations tend to quickly encounter compounding imitation error. 
    
    \textbf{High capacity models make the most use of unlabeled data.} In Fig.~\ref{fig:sweep_capacity} and Appendix~\ref{sec:appendix:sweep_capacity}, we additionally find the phenomenon that performance scales linearly with model size for policies that leverage unstructured data, whereas performance peaks and then declines for models trained on conventional predefined demonstrations. We see this holds even when datasets are restricted to the same size. This suggests that the simple recipe of collecting large unstructured imitation datasets, pairing them with a small amount of language data, then training large capacity imitation learning models may be a valid way to scale up language conditioned control.


    
    \begin{figure*}[htb]
      \centering
      \includegraphics[width=.9\linewidth]{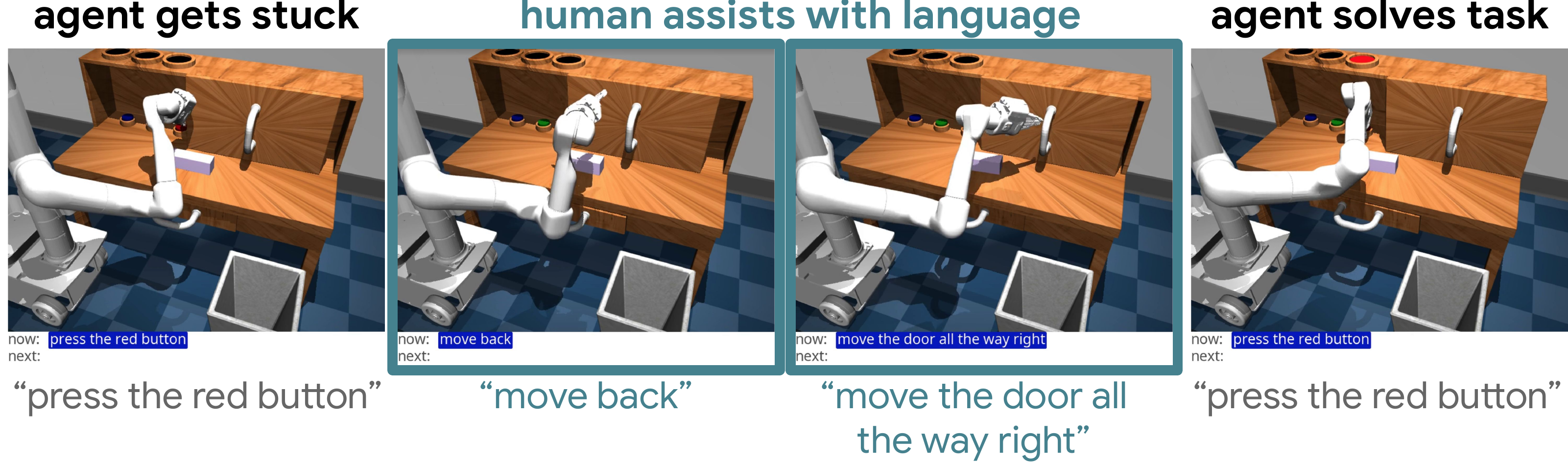}
      \caption{\textbf{
      Getting out of trouble with human language assistance:} 
      Unlike other forms of task conditioning, natural language conditioning allows a human operator to offer quick interactive assistance when an agent gets stuck. 
    }
    \label{fig:interaction}
    \end{figure*}

    
    \textbf{Language unlocks human assistance}.
    Natural language conditioning allows for new modes of interactive test time behavior, allowing humans to give guidance to agents that would be impractical to give via goal image or one-hot task conditioned control. See \fig{interaction}, \href{\website/\#interactive}{this video} and \sect{appendix:assistance} for a concrete example. Sec.~\ref{sec:qualitative_examples} additionally shows how humans can quickly compose tasks with language that are outside the 18-task benchmark (but covered by the training set), like ``put the block in the trash bin".

\subsection{Instruction Following with Large Pretrained Language Models}

    \begin{figure}[h!]
    \begin{minipage}{.5\textwidth}
        \includegraphics[width=1\linewidth]{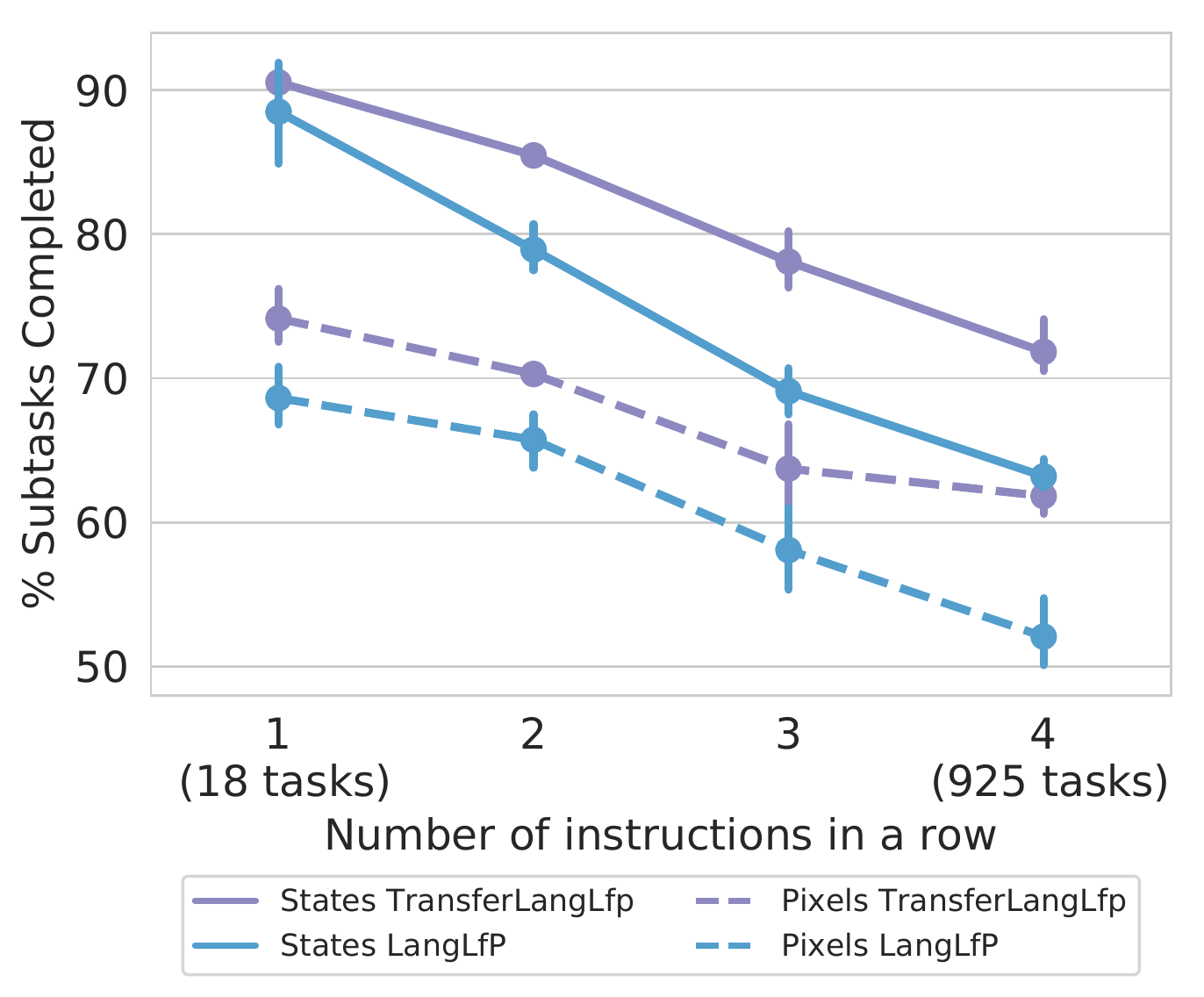}
        \caption{\textbf{Knowledge transfer from generic text corpora benefits robotic manipulation.} We see models that take pretrained embeddings as language input (purple) converge to higher performance than those that must learn language from scratch (blue).}
        \label{fig:systematic_transfer}
    \end{minipage}
    \end{figure}

    \textbf{Positive transfer to robotic manipulation}.
    In Table~\ref{tab:main_results} and \fig{systematic_transfer}, we see 
    that TransferLangLfP systematically outperforms LangLfP and LfP. This is the first evidence, to our knowledge, that sentence embeddings obtained from large pretrained language models can significantly improve the convergence of language-guided robotic control policies (Q3).

        
    \begin{table}[]
        {\centering
        {\small
        \begin{tabular}{|l|l|l|}
            \textbf{Method} & \textbf{\begin{tabular}[c]{@{}l@{}}OOD-syn\\ ($\sim$15k instructions)\end{tabular}} & \textbf{\begin{tabular}[c]{@{}l@{}}OOD-16-lang\\ ($\sim$240k instructions)\end{tabular}} \\ \hline
            Random Policy         & 0.0\% $\pm$ 0.0 & 0.0\% $\pm$ 0.0 \\
            LangLfP         & 37.6\% $\pm$ 2.3 & 27.94\% $\pm$ 3.5 \\
            TransferLangLfP       & \textbf{60.2\%} $\pm$ 3.2 & \textbf{56.0\%} $\pm$ 1.4
        \end{tabular}
        }
        
        \caption{\textbf{Out of distribution synonym robustness.}}
        \label{tab:ood}
        }
    \end{table}
        

\textbf{Robustness to out-of-distribution synonym instructions}.
In Table~\ref{tab:ood}, we study how robust our language conditioned policies are to synonyms not found in the training set. We see that only agents equipped with large pretrained language models (TransferLangLfP) are robust to these out of distribution synonyms (OOD-syn), giving affirmative support to Q4. Additionally, just by choosing a multilingual pretrained model \cite{yang2019multilingual}, we see this kind of robustness extends to 16 different languages, which have no vocabulary overlap with training (OOD-16-lang). See videos (\href{\website/\#multilingual}{link}) of TransferLangLfP following multilingual synonym instructions.
\textbf{We stress that these experiments do not test the ability of a policy to generalize to new \textit{kinds} of manipulation tasks than the ones seen in training}.  Rather, this shows that a simple training modification increases the number of \textit{ways} a language-guided agent can be conditioned to execute the same \textit{fixed set} of behaviors from its training distribution, effectively expanding the training instruction set. Find more details on these experiments in Appendix~\ref{sec:appendix:ood}.



\subsection{Limitations and Future Work}
    Although the coverage of unstructured play demonstration data mitigates failure modes in conventional imitation setups, we observe several limitations in our policies at test time.
    In this \href{\website/\#failure1}{video}, we see the policy make multiple attempts to solve the task, but times out before it is able to do so. We see in this \href{\website/\#failure2}{video} that the agent encounters a particular kind of compounding error, where the arm flips into an awkward configuration, likely avoided by humans during teleoperated play. This is potentially mitigated by a more stable choice of rotation representation, or more varied play collection. We note that the human is free to help the agent out of these awkward configurations using language assistance, as demonstrated in \sect{appendix:assistance}. More examples of failures can be seen \href{\website/\#failures}{here}.
    While LangLfP relaxes important constraints around task specification, it is fundamentally a goal-directed imitation method and lacks a mechanism for autonomous policy improvement. An exciting area for future work may be one that combines the coverage of teleoperated play, the scalability and flexibility of multicontext imitation pretraining, and the autonomous improvement of reinforcement learning, similar to prior successful combinations of LfP and RL \cite{gupta2019relay}.
    Additionally, the scope of this work is task agnostic control in a single simulated environment with a fixed set of objects. We note this is consistent with the standard imitation assumptions that training and test tasks are drawn i.i.d. from the same distribution. An interesting question for future work is whether training on a large play corpora covering many rooms and objects allows for generalization to unseen rooms or objects.

\section{Conclusion}
We proposed a scalable framework for combining multitask imitation with free-form text conditioning. Our method can learn language conditioned visuomotor policies, capable of following multiple human instructions over a long horizon in a dynamically accurate 3D tabletop setting. Key to our method is the ability to learn over unstructured and unlabeled imitation data, a property we made possible by introducing Multicontext Imitation Learning. Critically, the ability to learn from unstructured data reduced the cost of language annotation to less than 1\% of total data, while also resulting in much higher language conditioned task success. Finally, we showed a simple, but effective way to combine any language conditioned policy with large pretrained language models. We found that this small modification allowed our policy to be robust to many out-of-distribution synonym instructions, without requiring the collection of additional demonstration data.



\subsection*{Acknowledgments}
We thank Karol Hausman, Eric Jang, Mohi Khansari, Kanishka Rao, Jonathan Thompson, Luke Metz, Anelia Angelova, Sergey Levine, and Vincent Vanhoucke for providing helpful feedback on this manuscript. We additionally thank the annotators for providing paired language instructions.


\bibliographystyle{plainnat}
{\footnotesize
\bibliography{example}  
}

\newpage
\clearpage
\appendix
\section{Algorithm Pseudocode}
\subsection{Relabeling Play}
    \begin{algorithm}[h]
      \caption{Creating many goal image conditioned imitation examples from teleoperated play.}
      \label{alg:collect_play}
      \begin{algorithmic}[1]
        \STATE {\bfseries Input:} $\mathcal{S} = \{(s_{0:t}, a_{0:t})^n\}^{\infty }_{n}$, the unsegmented stream of observations and actions recorded during play.
        \STATE {\bfseries Input:} $\DPLAY \xleftarrow{} \{\}$.
        \STATE {\bfseries Input:} $w_{\operatorname{low}},\ w_{\operatorname{high}}$, bounds on hindsight window size.
        \WHILE{True}
        \STATE {\color{gray} \# Get next play episode from stream.}
        \STATE $(s_{0:t}, a_{0:t}) \thicksim \mathcal{S}$
        \FOR{$w = w_{\operatorname{low}} ... w_{\operatorname{high}}$}
          \FOR{$i = 0...(t-w)$}
              \STATE {\color{gray} \# Select each $w$-sized window.}
              \STATE $\tau = (s_{i:i+w}, a_{i:i+w})$
              \STATE {\color{gray} \# Treat last observation in window as goal.}
              \STATE $s_g = s_{w}$
              \STATE Add $(\tau, s_g)$ to $\DPLAY$
          \ENDFOR
        \ENDFOR
        \ENDWHILE
      \end{algorithmic}
    \end{algorithm}
    
\subsection{Pairing Play with Language}
    \begin{algorithm}[h]
      \caption{Pairing robot sensor data with natural language instructions.}
      \label{alg:collect_play_lang}
      \begin{algorithmic}[1]
      \STATE {\bfseries Input:} $\DPLAY$, a relabeled play dataset holding $(\tau, s_g)$ pairs.
      \STATE {\bfseries Input:} $\DPLAYLANG \xleftarrow{} \{\}$.
      \STATE {\bfseries Input:} $\operatorname{get\_hindsight\_instruction}()$: human overseer, providing after-the-fact natural language instructions for a given $\tau$.
      \STATE {\bfseries Input:} $K$, number of pairs to generate, $K << |\DPLAY|$.
      \FOR{$0 ... K$}
          \STATE {\color{gray} \# Sample random trajectory from play.}
          \STATE $(\tau, \_) \thicksim \DPLAY$
          \STATE {\color{gray} \# Ask human for instruction making $\tau$ optimal.}
          \STATE $l$ = $\operatorname{get\_hindsight\_instruction}(\tau)$
          \STATE Add $(\tau,\ l)$ to $\DPLAYLANG$
      \ENDFOR
      \end{algorithmic}
    \end{algorithm}
    
\section{LangLfP Implementation Details}
    \label{sec:appendix_architecture}
    Below we describe the networks we use for perception, natural language understanding, and control---all trained end-to-end as a single neural network to maximize our multicontext imitation objective. We stress that this is only one implementation of possibly many for LangLfP, which is a general high-level framework 1) combining relabeled play, 2) play paired with language, and 3) multicontext imitation.

    \begin{figure*}[h!t]
      \includegraphics[width=\linewidth]{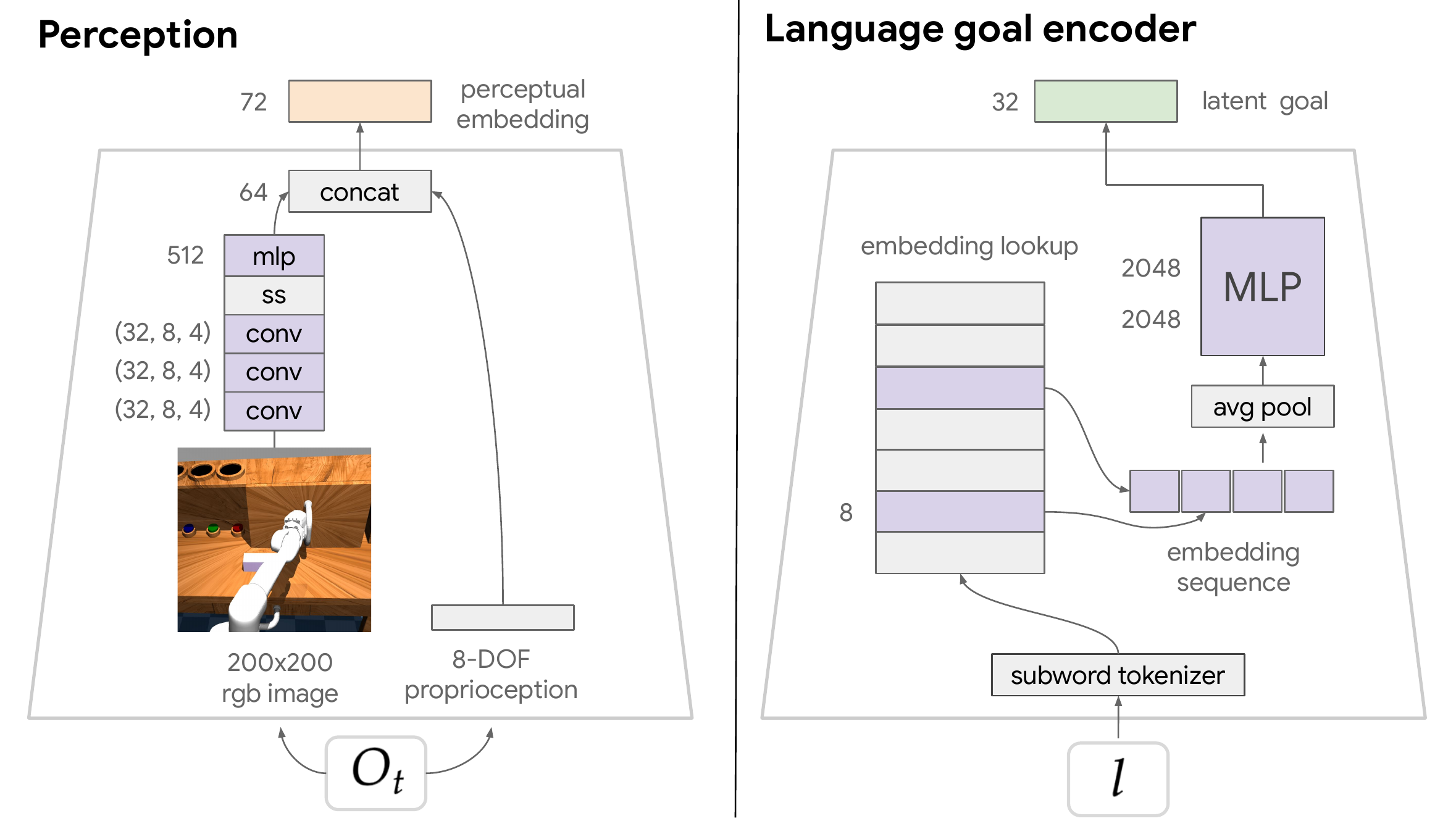}
      \caption{\textbf{Perception and language embedder modules}}
      \label{fig:perception_sentence_emb}
    \end{figure*}
    
\subsection{Perception Module}
    \label{sec:appendix_perception}
    We map raw observations (image and proprioception) to low dimensional perceptual embeddings that are fed to the rest of the network. To do this we use the perception module described in Fig.~\ref{fig:perception_sentence_emb}. We pass the image through the network described in Table~\ref{tab:vision_net_hparams}, obtaining a 64-dimensional visual embedding. We then concatenate this with the proprioception observation (normalized to zero mean, unit variance using training statistics). This results in a combined perceptual embedding of size 72.
    This perception module is trained end to end to maximize the final imitation loss. We do no photometric augmentation on the image inputs, but anticipate this may lead to better performance.
    \begin{table}
    \centering
    \begin{tabular}{ll}
    \hline
    Layer           & Details                     \\ \hline
    Input RGB Image & 200 x 200 x 3               \\
    Conv2D          & 32 filters, 8 x 8, stride 4 \\
    Conv2D          & 64 filters, 4 x 4, stride 2 \\
    Conv2D          & 64 filters, 3 x 3, stride 1 \\
    Spatial softmax &                             \\
    Flatten         &                             \\
    Fully connected & 512 hidden units, relu      \\
    Fully connected & 64 output units, linear     \\ \hline
    \end{tabular}
    \caption{Hyperparameters for vision network.}
    \label{tab:vision_net_hparams}
    \end{table}
    
\subsection{Image goal encoder}
\label{sec:appendix_genc}
$\genc$ takes as input image goal observation $O_g$ and outputs latent goal $z$. To implement this, we reuse the perception module $P_{\theta}$, described in Appendix~\ref{sec:appendix_perception}, which maps $O_g$ to $s_g$. We then pass $s_g$ through a 2 layer 2048 unit ReLU MLP to obtain $z$.

\subsection{Language understanding module}
    \label{sec:appendix_nlu}
    \textbf{From scratch}. For LangLfP, our latent language goal encoder, $\senc$, is described in Fig.~\ref{fig:perception_sentence_emb}. It maps raw text to a 32 dimensional embedding in goal space as follows: 1) apply subword tokenization, 2) retrieve learned 8-dim subword embeddings from a lookup table, 3) summarize the sequence of subword embeddings (we considered average pooling and RNN mechanisms for this, choosing average pooling based on validation), and 4) pass the summary through a 2 layer 2048 unit ReLU MLP. Out-of-distribution subwords are initialized as random embeddings at test time.
    
    \textbf{Transfer learning}. 
    For our experiments, we chose Multilingual Universal Sentence Encoder described in \cite{yang2019multilingual}. MUSE is a multitask language architecture trained on generic multilingual corpora (e.g. Wikipedia, mined question-answer pair datasets), and has a vocabulary of 200,000 subwords. MUSE maps sentences of any length to a 512-dim vector. We simply treat these embeddings as language observations and do not finetune the weights of the model. This vector is fed to a 2 layer 2048 unit ReLU MLP, projecting to latent goal space.
    
    We note there are many choices available for encoding raw text in a semantic pretrained vector space. MUSE showed results immediately, allowing us to focus on other parts of the problem setting. We look forward to experimenting with different choices of pretrained embedder in the future.
    
\subsection{Control Module}
    We could in principle use any goal-directed imitation network architecture to implement the policy $\pi_{\theta}(a_t | s_t, z)$. For direct comparison to prior work, we use Latent Motor Plans (LMP) architecture \citet{lynch2019play} to implement this network. LMP uses latent variables to model the large amount of multimodality inherent in freeform imitation datasets. Concretely it is a sequence-to-sequence conditional variational autoencoder (seq2seq CVAE), autoencoding contextual demonstrations through a latent ``plan" space. The decoder is a goal and latent plan conditioned policy. As a CVAE, LMP lower bounds maximum likelihood contextual imitation ($\mathcal{L}_{\operatorname{LfP}}$), and is easily adapted to our multicontext setting. 

    \label{sec:appendix_control}
    \textbf{Multicontext LMP}. Here we describe ``multicontext LMP": LMP adapted to be able to take either image or language goals. This imitation architecture learns both an abstract visuo-lingual goal space $z^g$, as well as a plan space $z^p$ capturing the many ways to reach a particular goal. We describe this implementation now in detail.

    As a conditional seq2seq VAE, original LMP trains 3 networks. 1) A posterior $q(z^p|\tau)$, mapping from full state-action demonstration $\tau$ to a distribution over recognized plans. 2) A learned prior $p(z^p|s_0, s_g)$, mapping from initial state in the demonstration and goal to a distribution over possible plans for reaching the goal. 3) A goal and plan conditioned policy $p(a_t|s_t, s_g, z^p)$, decoding the recognized plan with teacher forcing to reconstruct actions that reach the goal. See \cite{lynch2019play} for details.
 
    To train multicontext LMP, we simply replace the goal conditioning on $s_g$ everywhere with conditioning on $z^g$, the latent goal output by multicontext encoders $\mathcal{F}= \{\genc,\senc\}$. In our experiments, $\genc$ is a 2 layer 2048 unit ReLU MLP, mapping from encoded goal image $s_g$ to a 32 dimensional latent goal embedding. $\senc$ is a subword embedding summarizer described in Sec.~\ref{sec:appendix_nlu}. Unless specified otherwise, the LMP implementation in this paper uses the same hyperparameters and network architecture as in \cite{lynch2019play}. See appendix there for details on the posterior, conditional prior, and decoder architecture, consisting of a RNN, MLP, and RNN respectively.
    
\subsection{Training Details}
    \label{sec:appendix_training}
    At each training step, we compute two contextual imitation losses: image goal and language goal. The image goal forward pass is described in Fig.~\ref{fig:im_goal_lmp}. The language goal forward pass is described in Fig.~\ref{fig:lang_goal_lmp}. We share the perception network and LMP networks (posterior, prior, and policy) across both steps. We average minibatch gradients from image goal and language goal passes and we train everything---perception networks, $\genc$, $\senc$, posterior, prior, and policy---end-to-end as a single neural network to maximize the combined training objective. We describe all training hyperparameters in Table~\ref{tab:training_hparams}.

    \begin{table*}
    \centering
    \begin{tabular}{l|l}
    \hline
    \textbf{Hyperparameter} & \textbf{Value} \\ \hline
    hardware configuration & 8 NVIDIA V100 GPUs \\
    action distribution & discretized logistic, 256 bins per action dimension \\
    optimizer & Adam \\
    learning rate & 2e-4 \\
    hindsight window low & 16 \\
    hindsight window high & 32 \\
    LMP $\beta$ & 0.01 \\
    batch size (per GPU) & 64 sequences * padded max length 32 = 2048 frames \\
    training time & 3 days \\ \hline
    \end{tabular}
    \caption{\textbf{Training Hyperparameters}}
      \label{tab:training_hparams}
    \end{table*}
    
\subsection{Test time details}
    \label{sec:appendix_test}
    At the beginning of a test episode the agent receives as input its onboard observation $O_t$ and a human-specified natural language goal $l$. The agent encodes $l$ in latent goal space $z$ using the trained sentence encoder $\senc$. The agent then solves for the goal in closed loop, repeatedly feeding the current observation and goal to the learned policy $\pi_{\theta}(a|s_t, z)$, sampling actions, and executing them in the environment. The human operator can type a new language goal $l$ at any time.
    
\section{Environment}
    \label{sec:appendix_env}
    We use the same simulated 3D playground environment as in \cite{lynch2019play}, keeping the same observation and action spaces. We define these below for completeness.
    
    \begin{figure}
    \begin{center} \includegraphics[width=.8\linewidth]{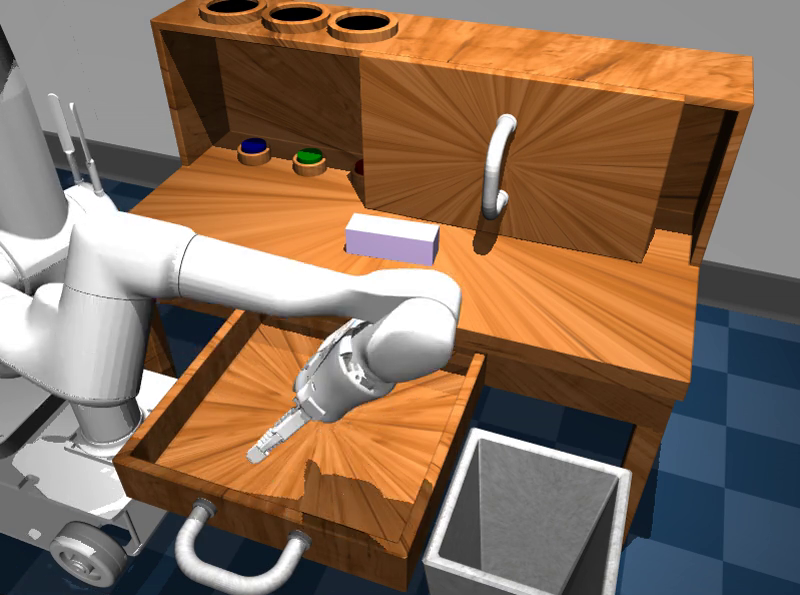}
    \caption{\textbf{The Playground environment.}}
    \label{fig:env}
    \end{center}
    \end{figure}
    
    \subsection{3D simulated environment}
    The environment contains a desk with a sliding door and drawer that can be opened and closed. On the desk is a movable rectangular block and 3 buttons that can be pushed to control different colored lights. Next to the desk is a trash bin. Physics are simulated using the MuJoCo physics engine \cite{TodorovET12}. Videos \href{\lfpvideos/play_data516x360.mp4}{here} of play data collected in this environment give a good idea of the available complexity. In front of the desk is a realistic simulated 8-DOF robot (arm and parallel gripper). The agent perceives its surroundings from egocentric high-dimensional RGB video sensors. It additionally has access to continuous internal proprioceptive sensors, relaying the cartesian position and rotation of its end effector and gripper angles. We modify the environment to include a text channel which the agent observes at each timestep, allowing humans to type unconstrained language commands. The agent must perform high-frequency, closed-loop continuous control to solve for user-described manipulation tasks, sending continuous position, rotation, and gripper commands to its arm at 30hz.     

    \subsection{Observation space}
    We consider two types of experiments: pixel and state experiments. In the pixel experiments, observations consist of (image, proprioception) pairs of 200x200x3 RGB images and 8-DOF internal proprioceptive state. Proprioceptive state consists of 3D cartesian position and 3D euler orientation of the end effector, as well as 2 gripper angles.
    In the state experiments, observations consist of the 8D proprioceptive state, the position and euler angle
    orientation of a movable block, and a continuous 1-d sensor describing: door open amount, drawer open amount, red
    button pushed amount, blue button pushed amount, green button pushed amount. In both experiments, the agent additionally observes the raw string value of a natural language text channel at each timestep.

    \subsection{Action space}
    We use the same action space as \cite{lynch2019play}: 8-DOF cartesian position, euler, and gripper angle of the end effector. Similarly, during training we quantize each action element into 256 bins. All stochastic policy outputs are represented as discretized logistic distributions over quantization bins. 
    
\section{Datasets}
    \label{sec:appendix_dataset}
    
    \subsection{Play dataset} We use the same play logs collected in \cite{lynch2019play} as the basis for all relabeled goal image conditioned learning. This consists of $\thicksim$7h of play relabeled in to $\DPLAY$: $\thicksim$10M short-shorizon windows, each spanning 1-2 seconds.
    
    \subsection{(Play, language) dataset}
    \label{sec:appendix_dataset_hip}
    
    We pair 10K windows from $\DPLAY$ with natural language hindsight instructions using the interface shown in \fig{ui} to obtain $\DPLAYLANG$. See real examples below in Table~\ref{tab:ex_train_lang}. Note the language collected has significant diversity in length and content.
    
    \begin{figure}[h!]
      \centering
      \includegraphics[width=.5\linewidth]{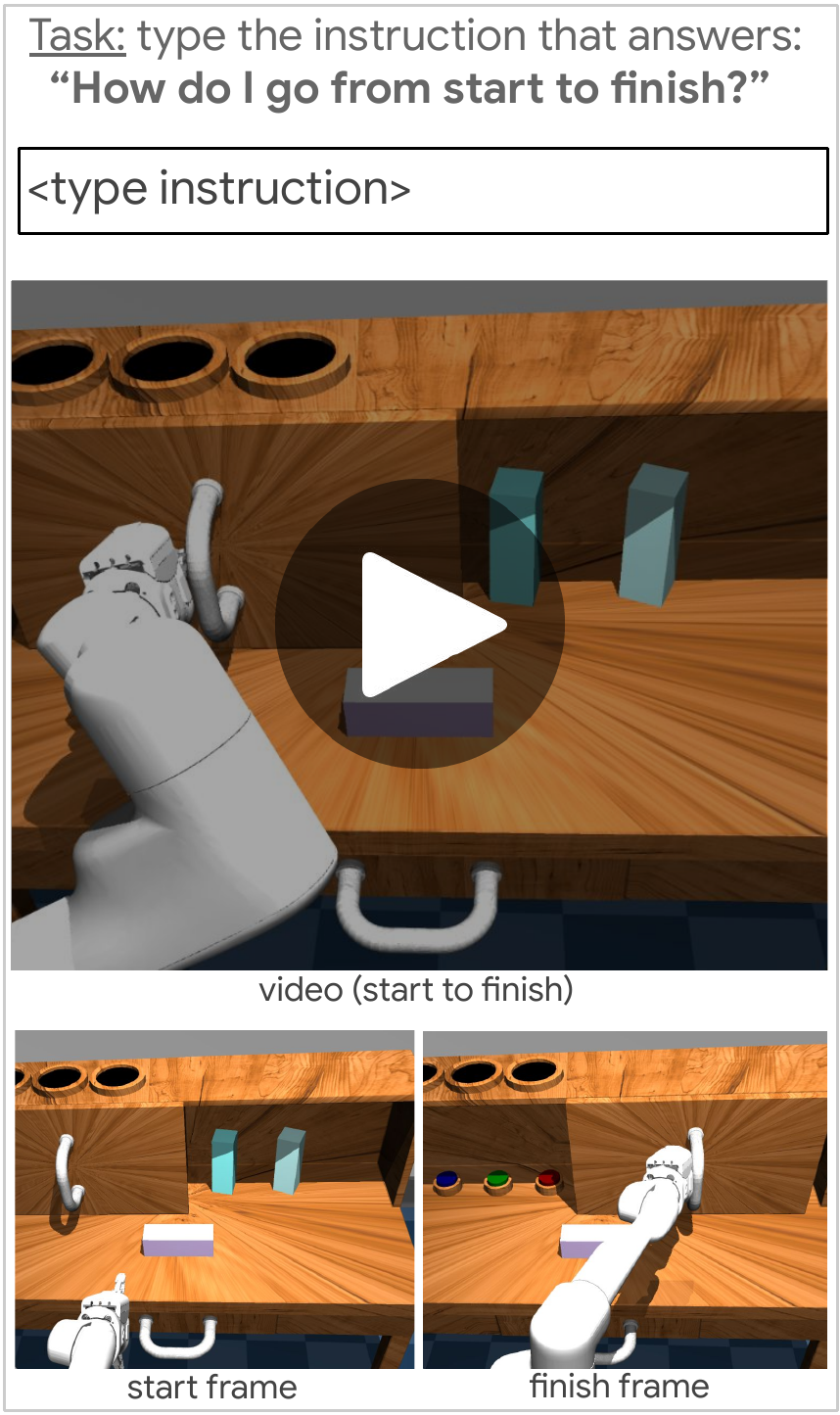}
      \caption{A schematic rendering of our hindsight language collection tool. }
      \label{fig:ui}
    \end{figure}
    
    \fig{ui} demonstrates how hindsight instructions are collected. Overseers are presented with a webpage that contains a looping video clip of 1 to 2 seconds of play, along with the first and the last frames of that video, to help them identify the beginning and the end of the clip if necessary.
    Overseers are asked to type in a text box the sentence that best answers the question "How do I go from start to finish?".
    Several considerations can go into the design of this interface. First, we can ask overseers to type in multiple instructions that are as different from each other as possible, to increase generalization and diversity of the language dataset. After experimenting with one or multiple text boxes, we opted for using only one, while asking users to write as diverse sentences as possible throughout the collection process. A disadvantage of having multiple boxes is that it can sometimes be challenging to come up with diverse sentences when the observed action is simple. It also leads to less video-language pairs for the same budget. Thus we decided one sentence per video was most beneficial in our case.
    Another collection consideration is the level of details in a sentence. For example, for a generalist robot application, it seems "open the drawer" is a more likely use case than "move your hand 10 cm to the right above the drawer handle, then grasp the drawer handle, then pull on the drawer handle". Similarly, an instruction geared toward a function such as "open the drawer" is more likely useful than one detached from it function, e.g. "put your fingers around the drawer handle, pull your hand back". Finally, given the temporal horizon of our video clips, multiple things can be happening within one clip. How many events should be described? For example it might be important to describe multiple motions such as "close the drawer then reach for the object". With all these observations in mind, we instructed the overseers to only describe the main actions, while asking themselves: “which instructions would I give to another human being so that they could produce a similar video if they were in that scene?”.
    
    \begin{table}
    \centering
    \begin{tabular}{|l|}
    \hline
    \begin{tabular}[c]{@{}l@{}}``grasp the object from the drawer and drop it\\on the table"\\ ``pick the object and then lift it up."\\ ``pull the drawer."\\ ``drag the object into the drawer"\\ ``drop the object,and again pickup the object high"\\ ``close the drawer"\\ ``do nothing." \\ ``move your hand to the right"\\ ``push the door to the right."\\ ``reach for the green button"\\ ``slightly close your hand." \\ ``go grasp the door handle"\\ ``close the drawer"\\ ``drop the object and push it inside the shelf"\\ ``grasp the block and drop it on top of the table"\\ ``place the block on top of the table"\\ ``rotate the block 90 degrees to the right"\\ ``move the cabinet door to the right and then release\\your fingers."\\ ``open the drawer"\\ ``slightly close the drawer, then drag it a little bit ,\\close it and then let go"\\ ``reach for the object."\\ ``press the red button and then release it"\\ ``move your hand towards the door handle"\\ ``drop the object"\\ ``slightly pull the drawer and then reach for its handle."\\ ``drop the object in the drawer"\\ ``release the block and move your hand upwards."\\ ``drop the object on the table and reach for\\the door handle"\\ ``push the object into to the drawer"\\ ``slightly move the door to the right, then drop the\\object and again turn the object to the left"\\ ``slowly move the door to the left"\end{tabular} \\ \hline
    \end{tabular}
    \caption{Randomly sampled examples from the training set of hindsight instructions paired with play. As hindsight instruction pairing sits on top of play, the language we collect is similarly not constrained by predefined tasks, and instead covers both specific functional behaviors and general, non task-specific behaviors.}
    \label{tab:ex_train_lang}
    \end{table}
    
    \begin{table}
    \centering
    \small{
    \begin{tabular}{l|l}
    \hline
    \textbf{Task}      & \textbf{Natural language instructions}         
    \\ \hline
    \begin{tabular}[c]{@{}l@{}}open sliding\\door\end{tabular}  & \begin{tabular}[c]{@{}l@{}}``move the door all the way to the right"\\ ``slide the door to the right"\\ ``move the sliding door all the way to the right\\and let go"\end{tabular} \\ \hline
    \begin{tabular}[c]{@{}l@{}}close sliding\\door\end{tabular} & \begin{tabular}[c]{@{}l@{}}``Grasp the door handle, then slide the door\\to the left"\\ ``move the door all the way to the left"\end{tabular}                                     \\ \hline
    open drawer        & \begin{tabular}[c]{@{}l@{}}```open the cabinet drawer"\\ ``open the drawer and let go"\end{tabular}                                                                               \\ \hline
    close drawer       & \begin{tabular}[c]{@{}l@{}}```close the drawer and let go"\\ ```close the drawer"\end{tabular}                                                                                    \\ \hline
    grasp flat         & \begin{tabular}[c]{@{}l@{}}``Pick up the block"\\ ``grasp the object and lift it up"\\ ``grasp the object and move your hand up"\end{tabular}                                     \\ \hline
    grasp lift         & \begin{tabular}[c]{@{}l@{}}``Pick up the object from the drawer\\and drop it on the table."\\ ``hold the block and place it on top of the table"\end{tabular}                      \\ \hline
    grasp upright      & \begin{tabular}[c]{@{}l@{}}``Pick the object and lift it up"\\ ``grasp the object and lift"\end{tabular}                                                                          \\ \hline
    knock              & \begin{tabular}[c]{@{}l@{}}``push the block forward"\\ ``push the object towards the door"\end{tabular}                                                                           \\ \hline
    pull out shelf     & \begin{tabular}[c]{@{}l@{}}``Drag the block from the shelf\\towards the drawer"\\ ``pick up the object from the shelf\\and drop it on the table"\end{tabular}                       \\ \hline
    put in shelf       & \begin{tabular}[c]{@{}l@{}}``grasp the object and place it inside\\the cabinet shelf"\\ ``Pick the object, move it into the shelf\\and then drop it."\end{tabular}                  \\ \hline
    push red           & \begin{tabular}[c]{@{}l@{}}``go press the red button"\\ ``Press the red button"\end{tabular}                                                                                      \\ \hline
    push green         & \begin{tabular}[c]{@{}l@{}}``press the green button"\\ ``push the green button"\end{tabular}                                                                                      \\ \hline
    push blue          & \begin{tabular}[c]{@{}l@{}}``push the blue button"\\ ``press down on the blue button."\end{tabular}                                                                               \\ \hline
    rotate left        & \begin{tabular}[c]{@{}l@{}}``Pick up the object, rotate it 90 degrees\\to the left and\\drop it on the table"\\ ``rotate the object 90 degrees to the left"\end{tabular}            \\ \hline
    rotate right       & \begin{tabular}[c]{@{}l@{}}``turn the object to the right"\\ ``rotate the block90 degrees to the right"\end{tabular}                                                              \\ \hline
    sweep              & \begin{tabular}[c]{@{}l@{}}``roll the object into the drawer"\\ ``drag the block into the drawer"\end{tabular}                                                                    \\ \hline
    sweep left         & \begin{tabular}[c]{@{}l@{}}``Roll the block to the left"\\ ``close your fingers and roll the object\\to the left"\end{tabular}                                                     \\ \hline
    sweep right        & \begin{tabular}[c]{@{}l@{}}``roll the object to the right"\\ ``Push the block to the right."\end{tabular}                                                                         \\ \hline
    \end{tabular}
    }
    \caption{Example natural language instructions used to specify the 18 test-time visual manipulation tasks.}
    \label{tab:ex_test_lang}
    \end{table}
    
    \subsection{(Demo, language) dataset}
    To define $\DDEMOLANG$, we pair each of the 100 demonstrations of the 18 evaluation tasks from \cite{lynch2019play} with hindsight instructions using the same process as in ~\ref{sec:hip}. We similarly pair the 20 test time demonstrations of each of the 18 tasks. See examples for each of the tasks in Table~\ref{tab:ex_test_lang}.
    $74.3\%$ of the natural language instructions in the test dataset appear at least once in $\DPLAY$.
    $85.08\%$ of the instructions in $\DPLAYLANG$ never never appear in the test set.
    
    \subsection{Restricted play dataset}
    For a controlled comparison between LangLfP and LangBC, we train on a play dataset restricted to the same size as the aggregate multitask demonstration dataset ($\thicksim$1.5h). This was obtained by randomly subsampling the original play logs $\thicksim$7h to $\thicksim$1.5h before relabeling.
    
\section{Models}
    \label{sec:appendix_models}
    Below we describe our various baselines and their training sources. Note that for a controlled comparison, all methods are trained using the exact same architecture described in Sec.~\ref{sec:appendix_architecture}, differing only on the source of data.
    \begin{table}[H]
        \centering
        \begin{tabular}{l|l}
            Method & Training Data \\
            \hline
            LangBC & $\DDEMOLANG$ \\
            LfP (goal image) & $\DPLAY$ \\ 
            LangLfP Restricted & restricted $\DPLAY$, $\DPLAYLANG$ \\ 
            LangLfP (ours) & $\DPLAY$, $\DPLAYLANG$ \\ 
            TransferLangLfP (ours) & $\DPLAY$, $\DPLAYLANG$ \\ 
        \end{tabular}
        \caption{Methods and their training data sources. All baselines are trained to maximize the same generalized contextual imitation objective MCIL.}
        \label{tab:methods}
    \end{table}
    
\section{Long Horizon Evaluation}
    \label{sec:appendix_eval}
    \textbf{Task construction}. 
    We construct long-horizon evaluations by considering transitions between the 18 subtasks defined in \cite{lynch2019play}. These span multiple diverse families, e.g. opening and closing doors and drawers, pressing buttons, picking and placing a movable block, etc. For example, one of the 925 Chain-4 tasks may be ``open\_sliding, push\_blue, open\_drawer, sweep". We exclude as invalid any transitions that would result in instructions that would have necessarily already been satisfied, e.g. ``open\_drawer, press\_red, open\_drawer". 
    
    Note this multi-stage scenario subsumes the original Multi-18 in \citet{lynch2019play}, which can be seen as just testing the first stage. This allows for direct comparison to prior work.
    
    To allow natural language conditioning we pair 20 test demonstrations of each subtask with human instructions using the same process as in Sec.~\ref{sec:hip} (example sentences in \tabl{ex_test_lang}).
    
    Success is reported with confidence intervals over 3 seeded training runs.
    
    \textbf{Evaluation walkthrough}. 
    The long horizon evaluation happens as follows. For each of the N-stage tasks in a given benchmark, we start by resetting the scene to a neutral state. This is discussed more in the next section.
    Next, we sample a natural language instruction for each task in the N-stage sequence from a test set. Next, for each subtask in a row, we condition the agent on the current subtask instruction and rollout the policy. If at any timestep the agent successfully completes the current subtask (according to the environment-defined reward), we transition to the next subtask in the sequence (after a half second delay). This attempts to mimic the qualitative scenario where a human provides one instruction after another, queuing up the next instruction, then entering it only once the human has deemed the current subtask solved. If the agent does not complete a given subtask in 8 seconds, the entire episode ends in a timeout.
    We score each multi-stage rollout by the percent of subtasks completed, averaging over all multi-stage tasks in the benchmark to arrive at the final N-stage number.
    
    \textbf{The importance of neutral starting conditions in multitask evaluation}. When evaluating any context conditioned policy (goal-image, task id, natural language, etc.), a natural question that arises is: how much is the policy relying on the context to infer and solve the task? In \cite{lynch2019play}, evaluation began by resetting the simulator to the first state of a test demonstration, ensuring that the commanded task was valid. We find that under this reset scheme, the initial pose of the agent can in some cases become correlated with the task, e.g. the arm starts nearby the drawer for a drawer opening task. This is problematic, as
    it potentially reveals task information to the policy through the \textit{initial state}, rather than the context. 
    
    In this work, we instead reset the arm to a fixed neutral position at the beginning of every test episode. This ``neutral" initialization scheme is used to run all evaluations in this paper. We note this is a fairer, but significantly more difficult benchmark. Neutral initialization breaks correlation between initial state and task, forcing the agent to rely entirely on language to infer and solve the task.
    For completeness, we present evaluation curves for both LangLfP and LangBC under this and the prior possibly  ``correlated" initialization scheme in Fig.~\ref{fig:chains_non_neutral}. We see that when LangBC is allowed to start from the exact first state of a demonstration (a rigid real world assumption), it does well on the first task, but fails to transition to the other tasks, with performance degrading quickly after the first instruction (Chain-2, Chain-3, Chain-4). Play-trained LangLfP on the other hand, is far more robust to change in starting position, experiencing only a minor degradation in performance when switching from the correlated initialization to neutral.

    \begin{figure}[htb]
      \centering
      \includegraphics[width=.8\linewidth]{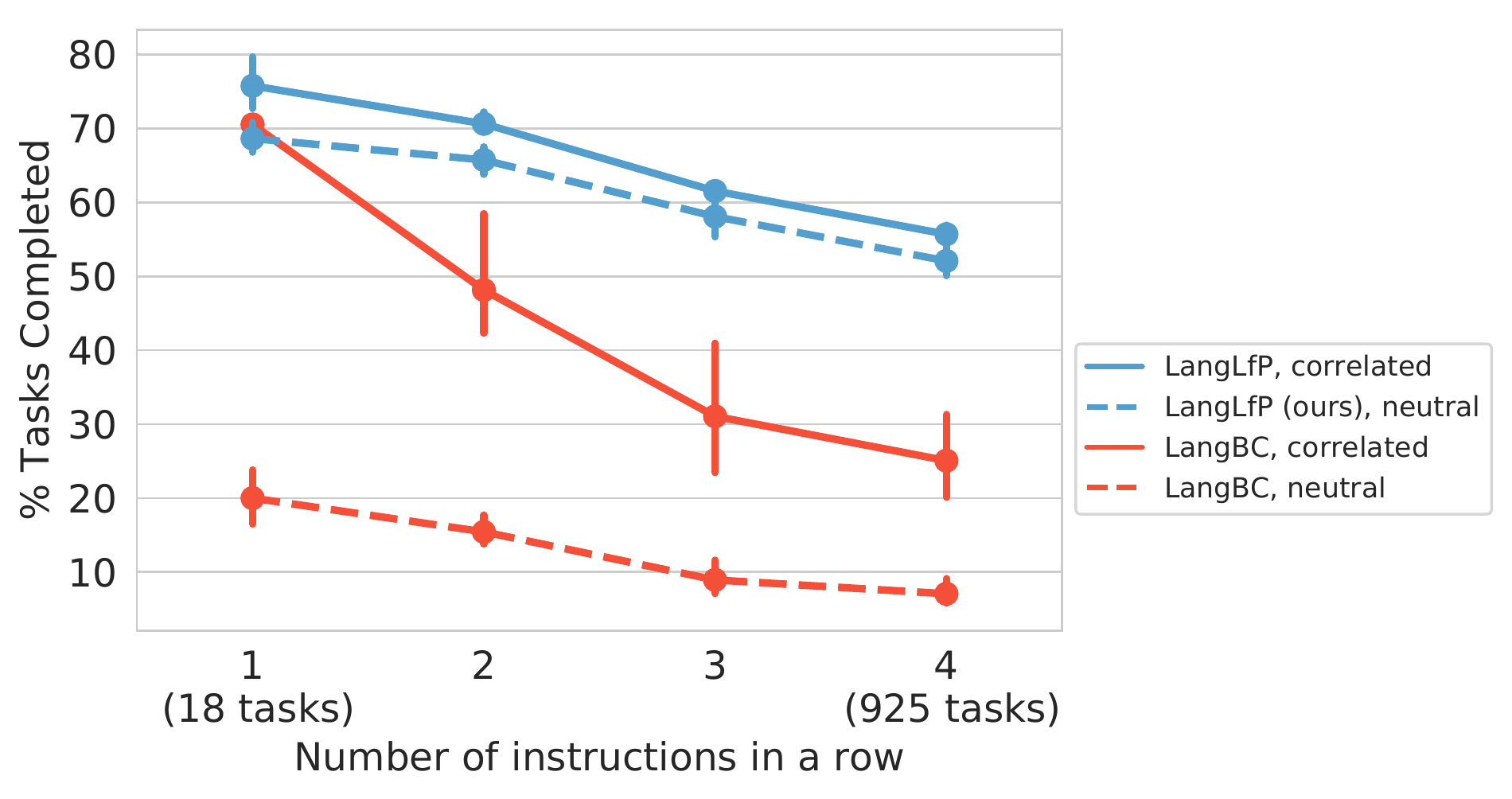}
      \caption{\textbf{Training on relabeled play leads to robustness.} Models trained on relabeled play (LangLfP) are robust to a fixed neutral starting position during multitask evaluation, models trained on conventional predefined demonstrations (LangBC) are not.}
      \label{fig:chains_non_neutral}
    \end{figure}

\section{Qualitative Examples}
\label{sec:qualitative_examples}

    \subsection{Composing new tasks with language}
    In this section, we show that LangLfP can achieve difficult new tasks, not included in the standard 18-task benchmark, in zero shot. In Fig.~\ref{fig:trash}, the operator can compose a new task ``put object in trash" at test time, simply by breaking the task into two text instructions: 1) ``pick up the object" and 2) ``put the object in the trash". We see similarly that the operator can ask the agent to put the object on top of a shelf in Fig.~\ref{fig:top_of_shelf}. Note that neither ``put on shelf" nor ``put in trash" are included in the 18-task benchmark.
    \begin{figure}[htb]
      \centering
      \includegraphics[width=.6\linewidth]{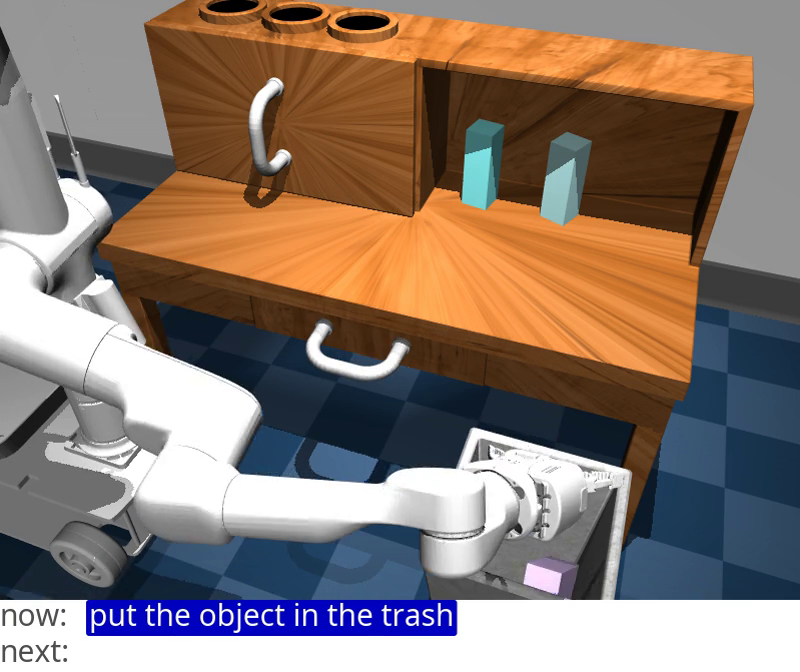}
      \caption{\textbf{Putting the object in the trash:} An operator is able to put the object in the trash by breaking down the task into 2 smaller subtasks with the following sentences; 1) ``pick up the object" 2) ``put the object in the trash"}
      \label{fig:trash}
    \end{figure}
    
    \begin{figure}[htb]
      \centering
      \includegraphics[width=.6\linewidth]{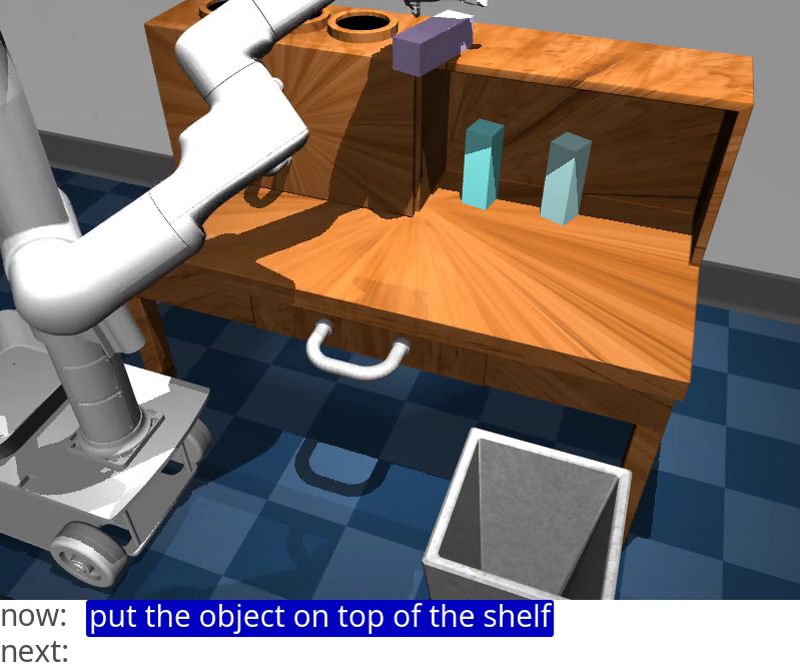}
      \caption{\textbf{Putting the object on top of the shelf:} An operator is able to put the object in the trash by breaking down the task into 2 smaller subtasks with the following sentences; 1) "pick up the object" 2) "put the object on top of the shelf".}
      \label{fig:top_of_shelf}
    \end{figure}
    
\subsection{Play scales with model capacity}
\label{sec:appendix:sweep_capacity}
In Fig.~\ref{fig:sweep_capacity}, we consider task performance as a function of model size for models trained on play or conventional predefined demonstrations. For fair comparison, we compare LangBC to Restricted LangLfP, trained on the same amount of data. We study both models from state inputs to understand upper bound performance. As we see, performance steadily improves as model size is increased for models trained on play, but peaks and then declines for models trained on conventional demonstrations. Our interpretation is that larger capacity is effectively utilized to absorb the diversity in play, whereas additional capacity is wasted on equivalently sized datasets constrained to predefined behaviors. This suggests that the simple recipe of collecting large play datasets and scaling up model capacity to achieve higher performance is a valid one.
    
\subsection{The Operator Can Help the Robot}
    \label{sec:appendix:assistance}
    
    In this section, we show an example of the human operator adapting its instructions if the robot gets stuck, and adding extra sub-steps to get out of trouble and fix the situation in order to achieve the initial instruction. In \fig{interaction}, the robot gets its end-effector stuck against the door on the way to pressing the red button. The operator then changing the command to move back, then opening the door more to the right so that it has better access to the red button, then pressing the red button.
    

\subsection{Pretrained language models give robustness to synonym instructions}
\label{sec:appendix:ood}
\textbf{Robustness to out of distribution synonym instructions}. To study whether our proposed transfer augmentation allows an agent to follow out-of-distribution synonyms, we replace one or more words in each Multi-18 \textbf{evaluation} instruction with a synonym outside training, e.g. ``drag the block from the shelf" $\xrightarrow{}$ ``\textit{retrieve} the \textit{brick} from the \textit{cupboard}". Note these substitutions only happen at test time, not training time. Enumerating all substitutions results in a set of 14,609 out-of-distribution instructions covering all 18 tasks. We evaluate a random policy, LangLfP, and TransferLangLfP on this benchmark, OOD-syn, reporting the results in Table~\ref{tab:ood}. Success is reported with confidence intervals over 3 seeded training runs. We see while LangLfP is able to solve some novel tasks by inferring meaning from context (e.g. 'pick up the block' and 'pick up the brick' might reasonably map to the same task for one freely moving object), its performance degrades significantly. TransferLangLfP on the other hand, degrades less. This shows that the simple transfer learning technique we demonstrate greatly magnifies the test time scope of an instruction following agent. We stress that this technique only increases the number of \textit{ways} a language-guided agent can be conditioned at test time to execute the same fixed set of training behaviors.

\textbf{Following out of distribution instructions in 16 different languages}. Here we show that simply choosing a multilingual pretrained language model leads to further robustness across languages not seen during training (e.g. French, Mandarin, German, etc.).
To study this, we combine the original set of \textbf{evaluation} instructions from Multi-18 with the expanded synonym instruction set OOD-syn, then translate both into 16 languages using the Google translate API. This results in $\thicksim$240k out-of-distribution instructions covering all 18 tasks. We evaluate the previous methods on this cross-lingual manipulation benchmark, OOD-16-lang, reporting success in Table~\ref{tab:ood}. We see that when LangLfP receives instructions with no training overlap, it resorts to producing maximum likelihood play actions. This results in some spurious task success, but performance degrades materially. TransferLangLfP, on the other hand, is far more robust to these perturbations, degrading only slightly from the english-only benchmark. See \href{\website/\#multilingual}{videos} of TransferLangLfP following instructions in 16 novel languages. While in this paper we do not finetune the pretrained embeddings with our imitation loss, an exciting direction for future work would be to see if grounding language embeddings in embodied imitation improves representation learning over text-only inputs.

\section{Ablation: How much language is necessary?}
    We study how the performance of LangLfP scales with the number of collected language pairs. Fig.~\ref{fig:lang_ablation} compares models trained from pixel inputs to models trained from state inputs as we sweep the amount of collected language pairs by factors of 2. Success is reported with confidence intervals over 3 seeded training runs. Interestingly, for models trained from states, doubling the size of the language dataset from 5k to 10k has marginal benefit on performance. Models trained from pixels, on the other hand, have yet to converge and may benefit from even larger datasets. This suggests that the role of larger language pair datasets is primarily to address the complicated perceptual grounding problem.
    
    \begin{figure}[htb]
      \centering
      \includegraphics[width=.6\linewidth]{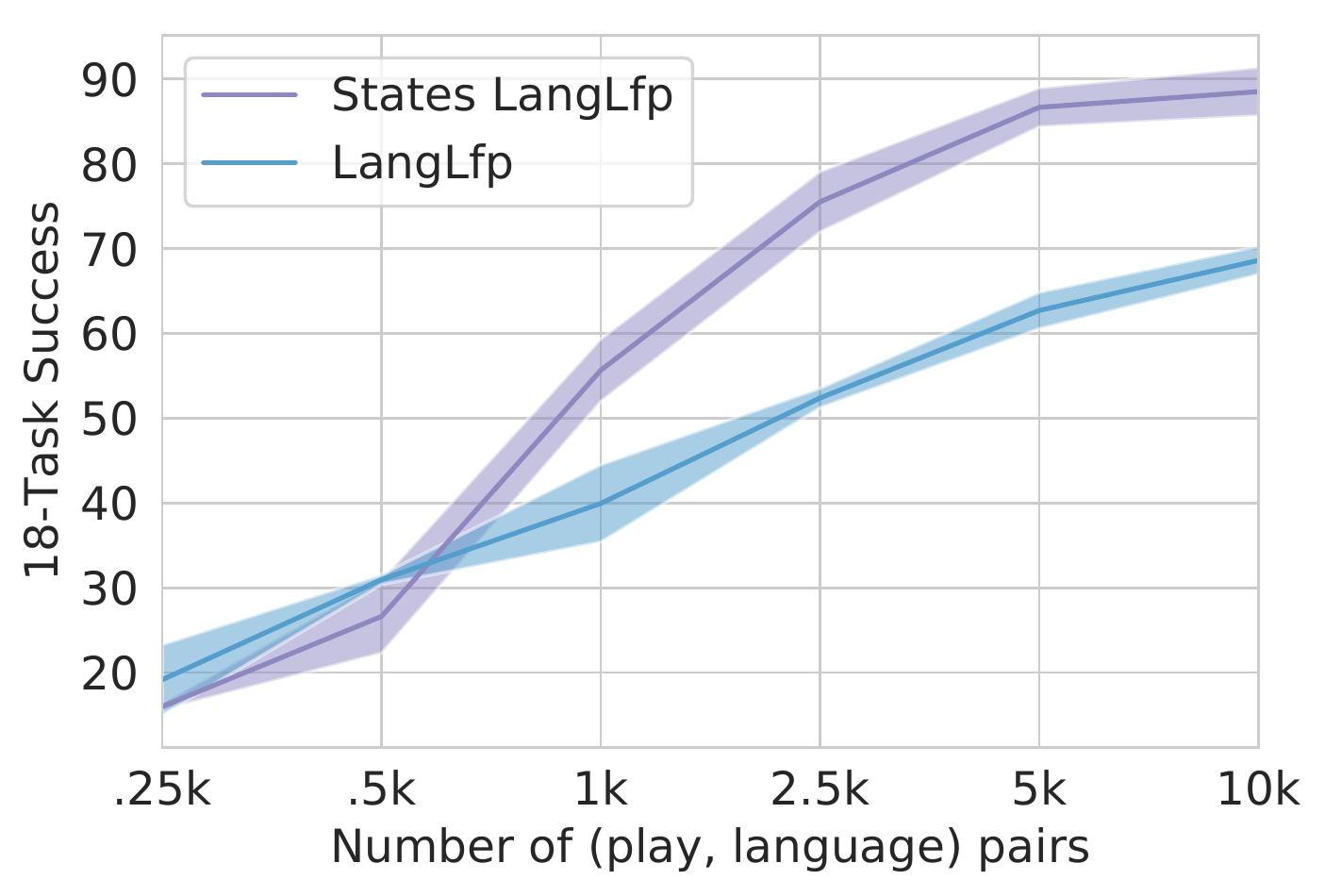}
      \caption{\textbf{18-task success vs amount of human language pairs collected.}}
      \label{fig:lang_ablation}
    \end{figure}
    
    \begin{figure*}[htb]
      \centering
      \includegraphics[width=1\linewidth]{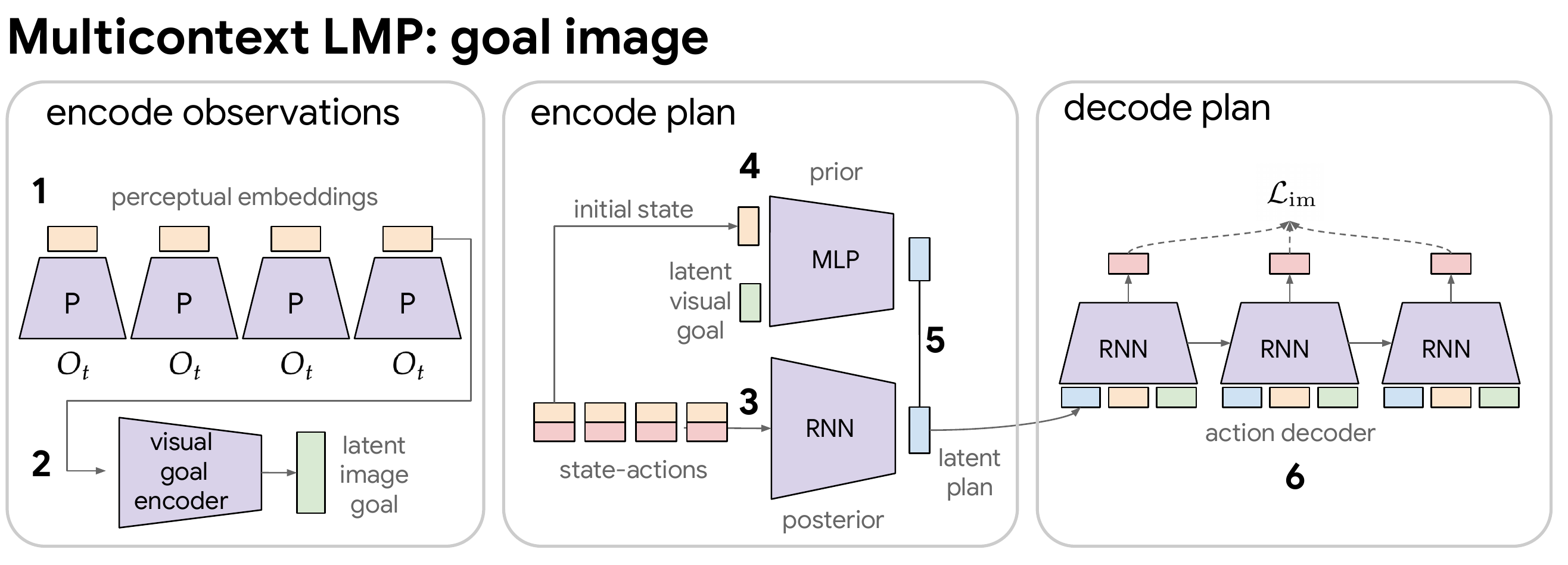}
      \caption{\textbf{Latent image goal LMP}: This image goal conditioned forward pass of multicontext LMP.
      1) Map sequence of raw observations $O_t$ to perceptual embeddings (see Fig.~\ref{fig:perception_sentence_emb}). 2) Map image goal to latent goal space. 3) Map full state-action sequence to recognized plan through posterior. 4) Map initial state and latent goal through prior network to distribution over plans for achieving goal. 5) Minimize KL divergence between posterior and prior. 6) Compute maximum likelihood action reconstruction loss by decoding plan into actions with teacher forcing. Each action prediction is conditioned on current perceptual embedding, latent image goal, and latent plan. Note perception net, posterior, prior, and policy are shared with language goal forward pass.}
      \label{fig:im_goal_lmp}
    \end{figure*}
    
    \begin{figure*}[htb]
      \includegraphics[width=\linewidth]{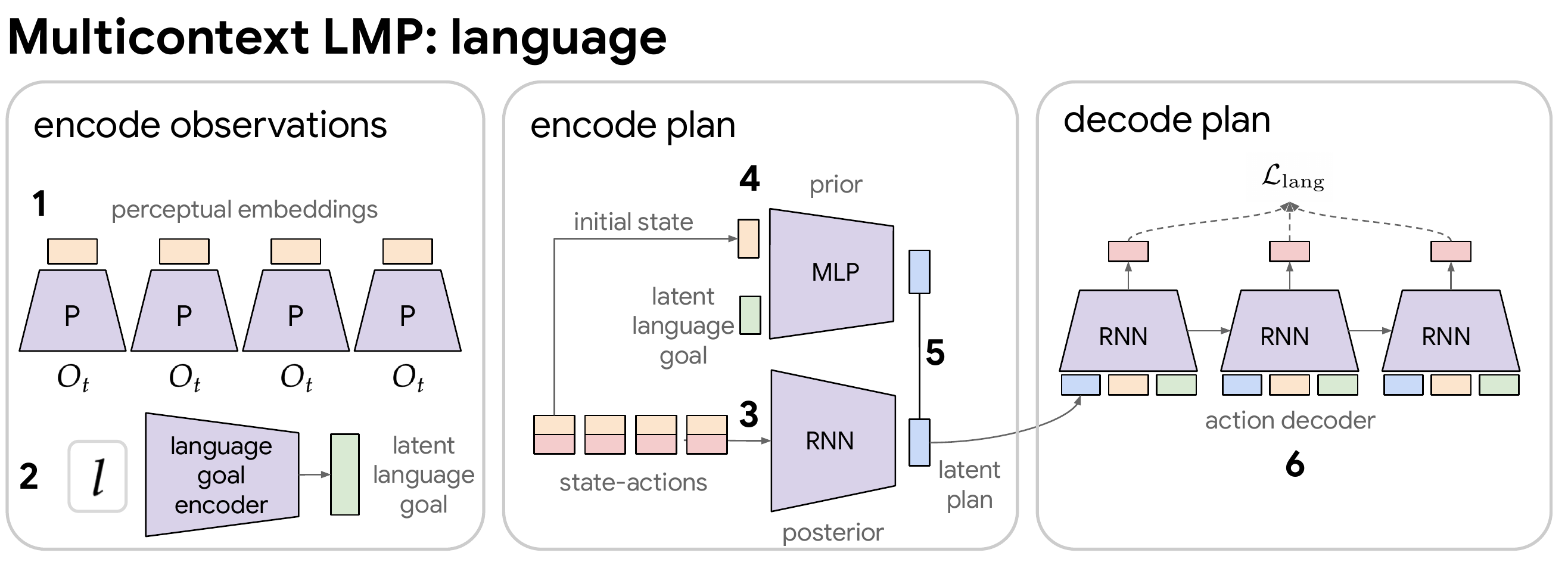}
      \caption{\textbf{Latent language goal LMP}: This describes the language goal conditioned forward pass of multicontext LMP. 1) Map sequence of raw observations $O_t$ to perceptual embeddings (see Fig.~\ref{fig:perception_sentence_emb}). 2) Map language observation to latent goal space. 3) Map full state-action sequence to recognized plan through posterior. 4) Map initial state and latent goal through prior network to distribution over plans for achieving goal. 5) Minimize KL divergence between posterior and prior. 6) Compute maximum likelihood action reconstruction loss by decoding plan into actions with teacher forcing. Each action prediction is conditioned on current perceptual embedding, latent language goal, and latent plan. Note perception net, posterior, prior, and policy are shared with image goal LMP.}
      \label{fig:lang_goal_lmp}
    \end{figure*}

\end{document}